\documentclass[10pt,twocolumn,letterpaper]{article}

\usepackage{iccv}
\usepackage{times}
\usepackage{epsfig}
\usepackage{graphicx}
\usepackage{amsmath}
\usepackage{amssymb}
\usepackage{booktabs}
\usepackage{multirow}
\usepackage{rotating}
\usepackage{tabularx}
\usepackage{multicol}
\usepackage[accsupp]{axessibility}

\usepackage[pagebackref=true,breaklinks=true,letterpaper=true,colorlinks,bookmarks=false]{hyperref}

\usepackage[capitalize]{cleveref}
\crefname{section}{Sec.}{Secs.}
\Crefname{section}{Section}{Sections}
\Crefname{table}{Table}{Tables}
\crefname{table}{Tab.}{Tabs.}

\iccvfinalcopy % *** Uncomment this line for the final submission

\def\iccvPaperID{5658} % *** Enter the ICCV Paper ID here

\ificcvfinal\fi

\begin{document}

%%%%%%%%% TITLE
\title{MixCycle: Mixup Assisted Semi-Supervised 3D Single Object Tracking with Cycle Consistency}

\author{
Qiao Wu$^{1}$\;
Jiaqi Yang$^{1}$\thanks{Corresponding author}\; 
Kun Sun$^{2}$\;
Chu'ai Zhang$^{1}$ \
Yanning Zhang$^{1}$ \;
Mathieu Salzmann$^{3}$ \\
$^1$  Northwestern Polytechnical University\;
$^2$  China University of Geosciences, Wuhan\\
$^3$ École Polytechnique Fédérale de Lausanne \\
\tt \small qiaowu@mail.nwpu.edu.cn; jqyang@nwpu.edu.cn; sunkun@cug.edu.cn;\\
\tt \small cazhang@mail.nwpu.edu.cn; ynzhang@nwpu.edu.cn; mathieu.salzmann@epfl.ch
}
% \and

\maketitle
\ificcvfinal\thispagestyle{empty}\fi

%%%%%%%%% ABSTRACT
\begin{abstract}
3D single object tracking (SOT) is an indispensable part of automated driving. Existing approaches rely heavily on large, densely labeled datasets. However, annotating point clouds is both costly and time-consuming. Inspired by the great success of cycle tracking in unsupervised 2D SOT, we introduce the first semi-supervised approach to 3D SOT. Specifically, we introduce two cycle-consistency strategies for supervision: 1) Self tracking cycles, which leverage labels to help the model converge better in the early stages of training; 2) forward-backward cycles, which strengthen the tracker's robustness to motion variations and the template noise caused by the template update strategy. Furthermore, 
we propose a data augmentation strategy named SOTMixup to improve the tracker's robustness to point cloud diversity. 
SOTMixup generates training samples by sampling points in two point clouds with a mixing rate and assigns a reasonable loss weight for training according to the mixing rate.
The resulting MixCycle approach generalizes to appearance matching-based trackers. On the KITTI benchmark, based on the P2B tracker~\cite{qi_p2b_2020}, MixCycle trained with $\textbf{10\%}$ labels outperforms P2B trained with $\textbf{100\%}$ labels, and achieves a $\textbf{28.4\%}$ precision improvement when using $\textbf{1\%}$ labels. Our code will be released at \url{https://github.com/Mumuqiao/MixCycle}.
\end{abstract}

%%%%%%%%% BODY TEXT
\section{Introduction}
\label{sec:intro}
3D single object tracking (SOT) plays a critical role in the field of autonomous driving. For example, given object detection~\cite{qi_deep_2019,zhao_sess_2020} results as input, it can output the necessary information for trajectory prediction~\cite{hazard_importance_2022}. The goal of SOT is to regress the center position and 3D bounding-box (BBox) of an object of interest in a search area, given the point cloud (PC) patch and BBox of the object template. This is a very challenging task because (i) point clouds obtained with, e.g., LiDAR sensors, suffer from occlusions and point sparsity, 
complicating the tracker's task of finding the object of interest; (ii) the point distribution for an object may vary significantly, making it difficult for the model to learn discriminative object features.

\begin{figure}[t]
  \centering
%   \fbox{\rule{0pt}{2in} \rule{0.9\linewidth}{0pt}}
   \includegraphics[width=1.0\linewidth]{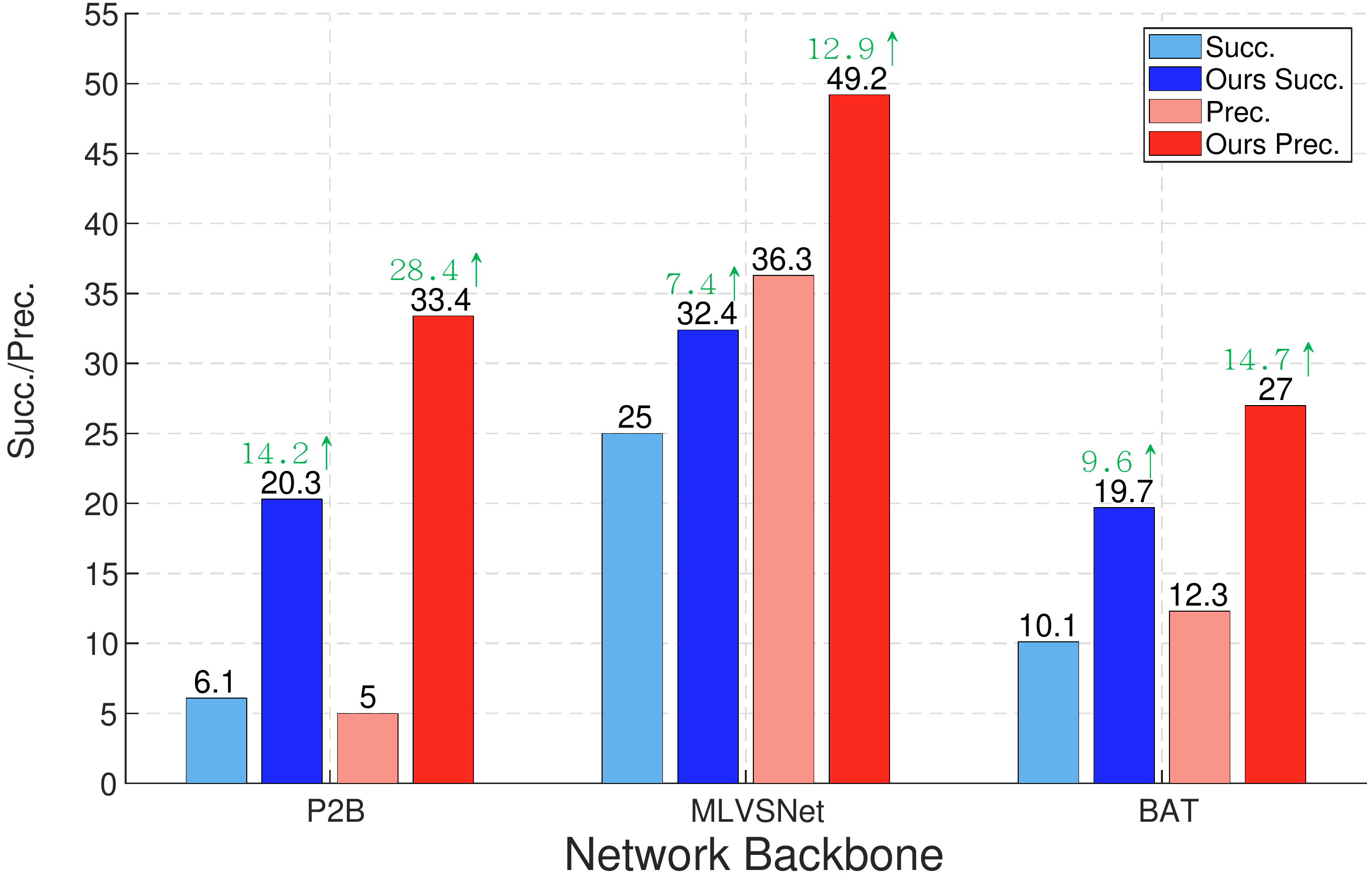}

   \caption{Comparison of MixCycle and fully-supervised methods~\cite{qi_p2b_2020,wang_mlvsnet_2021,zheng_box-aware_2021}, all trained with $1\%$ labels on KITTI~\cite{geiger_are_2012}. `Succ.' and `Prec.' represent Success and Precision, respectively.}
   \label{fig:resultKITTI}
\end{figure}

To tackle the above challenges, existing 3D SOT models~\cite{giancola_leveraging_2019,qi_p2b_2020,fang_3d-siamrpn_2020,hui_3d_2021,hui_3d_2022,wang_mlvsnet_2021,zheng_box-aware_2021,shan_ptt_2021,zheng_beyond_2022} rely on large scale annotated point cloud datasets for training. Unfortunately, obtaining annotations for this task, as for many 3D tasks, is extremely time-consuming. Furthermore, as shown in \cref{fig:resultKITTI}, the performance of these methods degrades dramatically as the number of labeled samples decreases. Nevertheless, no semi-supervised or unsupervised methods have been explored so far in 3D SOT.

\begin{figure}[tp]
  \centering
    \includegraphics[width=1.0\linewidth]{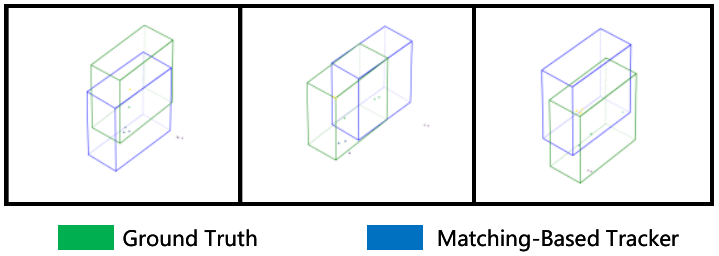}
    \caption{We observe that appearance matching-based trackers can learn the objects motion distribution and track them even in the absence of points for appearance matching. For instance, BAT~\cite{zheng_box-aware_2021} manages to track objects in  extremely sparse point clouds.}
  \label{fig:match-motion}
\end{figure}

%cycle Motivation
As shown in \cref{fig:match-motion}, the matching-based tracker of~\cite{zheng_box-aware_2021} can still track the target at the very beginning of a sequence by predicting a motion offset relative to the reference coordinate, even though there are no points in the template for appearance matching. This indicates that the appearance matching-based trackers can learn motion information.
By contrast, in 2D SOT, many trackers~\cite{wang_unsupervised_2019,wang_unsupervised_2021,zheng_learning_2021,yuan_self-supervised_2020} employ cycle consistency to leverage unsupervised data. Specifically, they encourage forward and backward tracking to produce consistent motions. In principle, we expect to apply this idea to 3D SOT and make trackers to learn the object's motion distribution in unlabeled data.
However, transferring these 2D methods directly to 3D is challenging.
First, since the point cloud is sparse and the environment is cluttered with objects, it is hard to find meaningful patches to use as pseudo labels for training.
Second, unsupervised 2D SOT methods rely on the assumption that the target appears in every frame of the sequence. Unfortunately, this assumption is not always satisfied in point cloud datasets such as KITTI~\cite{geiger_are_2012}, NuScenes~\cite{caesar_nuscenes_2019}, and Waymo~\cite{sun_scalability_2020}. This is because they are built for the multi-object tracking task, and cannot guarantee that the tracking object exists in the whole sequence.

%-------------------------------------------------------------------------------------

In this paper, we introduce a label-efficient way to train 3D SOT trackers. We call it \textbf{MixCycle} - a 3D SOT approach based on a novel SOT\textbf{Mix}up data augmentation strategy for semi-supervised \textbf{Cycle} tracking. %with transformation consistency 
Specifically, we first develop a tracking framework exploiting both self and forward-backward tracking cycles. Self tracking consistency is performed to cover the object point cloud appearance variation, and forward-backward consistency is built for learning the object's motion distribution.
Second, we present a data augmentation method for 3D SOT called SOTMixup, which is inspired by the success of \emph{mixup}~\cite{zhang_mixup_2018} and \emph{Manifold mixup}~\cite{verma_manifold_2019}.
Without changing the total number of points in the search area, SOTMixup samples points in a random point cloud and the search area point cloud according to the mixing rate and generates training samples.
Specifically, the random point cloud is sampled from the labeled training set. 
SOTMixup thus increases the tracker's robustness to point cloud variations.
We evaluate MixCycle on KITTI, NuScenes, and Waymo. As shown in~\cref{fig:resultKITTI}, our experiments clearly demonstrate the label efficiency, generalization and remarkable performance of our method on the 3D SOT task.
%-------------------------------------------------------------------------------------

\noindent\textbf{Contributions:} \textbf{(i)} We propose the first semi-supervised 3D SOT framework. It exploits self and forward-backward consistency as supervision and generalizes to appearance matching-based trackers. \textbf{(ii)} We introduce a SOTMixup augmentation strategy that increases the tracker's robustness to point distribution variations and allows it to learn motion information in extreme situations. \textbf{(iii)} Our framework demonstrates a remarkable performance in terms of label efficiency, achieving better results than existing supervised methods in our experiments on KITTI NuScenes, and Waymo when using fewer labels. In particular, we surpass P2B~\cite{qi_p2b_2020} trained on $100\%$ labels while only using $10\%$ labels.

%------------------------------------------------------------------------
\section{Related Work}
\label{sec:relate}
\noindent \textbf{3D Single Object Tracking.} Since LiDAR is insensitive to illumination, the appearance matching model has become the main choice in the field of 3D single object tracking. Giancola \etal~\cite{giancola_leveraging_2019} proposed SC3D which is the first method using a Siamese network to deal with this problem. However, it is very time-consuming and inaccurate due to heuristic matching. Zarzar \etal~\cite{zarzar_efficient_2019} built an end-to-end tracker by using 2D RPN in 2D bird's eyes view (BEV). Unfortunately, the lack of information in one dimension leads to limited accuracy. The point-to-box (P2B) network~\cite{qi_p2b_2020} employs VoteNet\cite{qi_deep_2019} as object regression module to construct a point-based tracker. A number of works~\cite{fang_3d-siamrpn_2020,hui_3d_2021,hui_3d_2022,wang_mlvsnet_2021,zheng_box-aware_2021,shan_ptt_2021} investigate different architectures of trackers based on P2B~\cite{qi_p2b_2020}. Zheng \etal~\cite{zheng_box-aware_2021} depicted an object using the point-to-box relation and proposed BoxCloud, which enables the model to better sense the size of objects. Hui \etal~\cite{hui_3d_2021} discovered the priori information of object shapes in the dataset to obtain dense representations of objects from sparse point clouds. Zheng \etal~\cite{zheng_beyond_2022} presented a motion centric method $M^2$-Track, which is appearance matching-free and has made great progress in dealing with the sparse point cloud tracking problem. However, $M^2$-track is limited by the LiDAR frequency of datasets as object motion in adjacent frames varies with the LiDAR sampling frequency.
%limited by the LiDAR frequency of datasets.

All the above methods rely on large-scale labeled datasets. Unfortunately, 3D point cloud annotation is labor- and time-consuming. To overcome this, we propose MixCycle, a semi-supervised tracking method based on cycle consistency constraints with SOTMixup data augmentation.

%-------------------------------------------------------------------------------------

\begin{figure*}[t]
  \centering
  \includegraphics[width=1.0\linewidth]{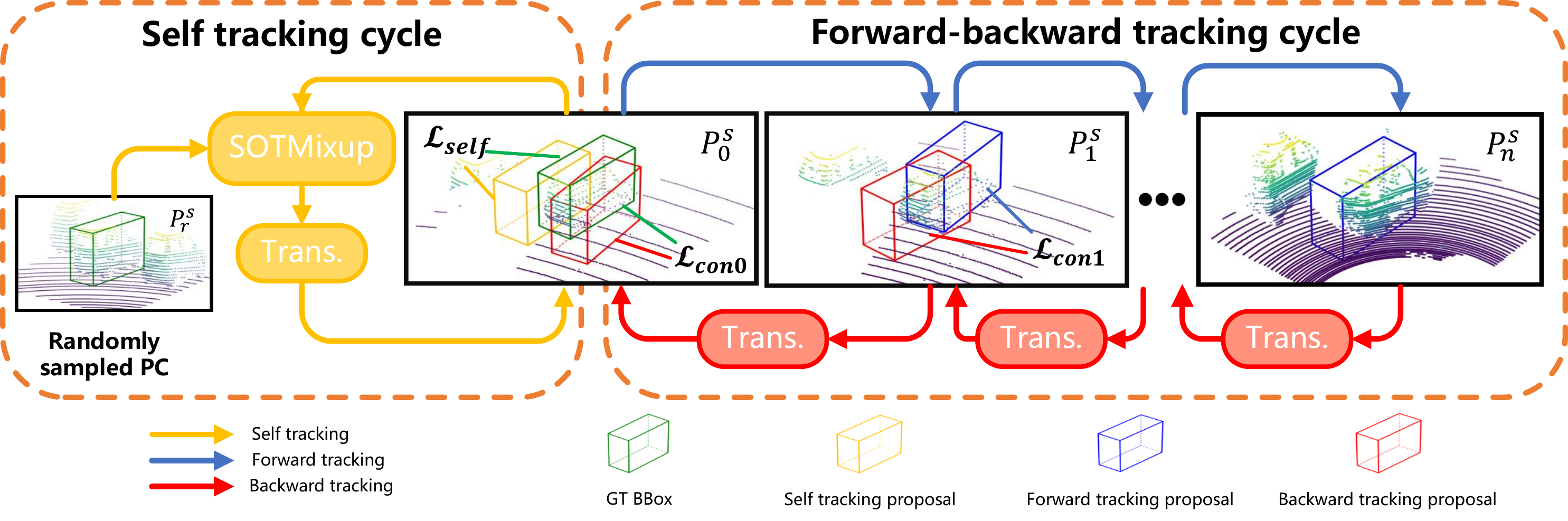}
    \caption{\textbf{MixCycle framework.} The label is only contained in $P^{s}_{0}$ of point cloud (PC) sequence $\{P^{s}_{0}, P^{s}_{1},\dots, P^{s}_{n}\}$. \textbf{1) Self tracking cycle:} we first sample a PC $P^{s}_{r}$ from the labeled training set. Then, we generate pseudo labels by applying SOTMixup and random rigid transformation (Trans.) to $P_{0}^{s}$ and $P^{s}_{r}$. SOTMixup directly mixes $P^{s}_{r}$ and $P_{0}^{s}$ based on the number of points with a mixing rate, assigning a reasonable loss weight corresponding to the mixing rate. We employ the consistency between self tracking proposals and pseudo labels to formulate the loss $\mathcal{L}_{self}$. \textbf{2) Forward-backward tracking cycle:} we leverage forward tracking proposals as pseudo labels and apply a random rigid transformation to them. Then, we employ the consistency between ground truth (GT)/pseudo labels and backward tracking proposals to formulate the losses $\mathcal{L}_{con0}$/$\{\mathcal{L}_{con1},\dots, \mathcal{L}_{con(n-1)}\}$.}
  \label{fig:pipline}
\end{figure*}

\noindent \textbf{Label-Efficient Visual Tracking.} Wang \etal~\cite{wang_unsupervised_2019} proposed unsupervised deep tracking (UDT) with cycle consistency, based on a Siamese correlation filter backbone network. UDT achieved remarkable performance, revealing the potential of unsupervised learning in visual tracking. Yuan \etal~\cite{yuan_self-supervised_2020} improved the UDT approach to make the target features passed forward and backward as similar as possible. The self-supervised fully convolutional Siamese network~\cite{sio_s2siamfc_2020} uses only spatially supervised learning of target correspondences in still video frames. Wu \etal~\cite{wu_progressive_2021} proposed a progressive unsupervised learning (PUL) network, which distinguishes the background by contrastive learning and models the regression result noise. PUL thus makes the tracker robust in long-time tracking. Unsupervised single object tracker~\cite{zheng_learning_2021} consists of an online-updating tracker with a novel memory learning scheme.

In essence, the above unsupervised trackers all make the implicit assumption that the tracked target exists in every frame of the sequence. 
Unfortunately, this is not necessarily true in KITTI~\cite{geiger_are_2012}, NuScenes~\cite{caesar_nuscenes_2019}, and Waymo~\cite{sun_scalability_2020}. 
Therefore, the above methods are not directly applicable to 3D SOT.

%-------------------------------------------------------------------------------------

\noindent \textbf{Mixup Data Augmentation.} Data augmentation has become a crucial pre-processing step for many deep learning models. Zhang \etal~\cite{zhang_mixup_2018} introduced a data augmentation method called \emph{mixup}, which linearly interpolates two image samples, and \emph{Manifold mixup} ~\cite{verma_manifold_2019} transfers this idea to high-dimensional feature spaces. By interpolating a new sample, PointMixup~\cite{chen_pointmixup_2020} extends \emph{mixup} to point clouds. Mix3D~\cite{nekrasov_mix3d_2021} introduces a scene-aware \emph{mixup} by taking the union of two 3D scenes and their labels after random transformations. Lu \etal~\cite{lu_directed_2022} developed a directed \emph{mixup} based on the pixel values. Additionally, a variety of region \emph{mixup} techniques have been proposed~\cite{zhang_pointcutmix_2022,lee_sagemix_2022,umam_point_2022,lee_regularization_2021}. In the case of outdoor scenes, Xiao \etal~\cite{xiao_polarmix_2022}  combined two images using random rotation. CosMix\cite{saltori_cosmix_2022} and structure aware fusion~\cite{hasecke_what_2022} combine point clouds using semantic structures. Fang \etal~\cite{fang_lidar-aug_2021} turned a CAD into a point cloud to combat object occlusion.

However, the above-mentioned methods are made for the multi-class classification scenario and are not suitable for SOT, which only contains a positive and negative sample.
%-------------------------------------------------------------------------

\section{Method}
\label{sec:method}

%-------------------------------------------------------------------------------------

\subsection{Overview}
\label{sec:overview}
The purpose of 3D SOT is to continually locate the target in the search area point cloud sequence $\mathbf{P^s} = \{P^{s}_{0},\dots, P^{s}_{k},\dots, P^{s}_{n} | P^{s}_{k} \in \mathbb{R}^{N_{s} \times 3}\}$ given the tracking object template point cloud $P^{o}_{0} \in \mathbb{R}^{N_{t} \times 3}$ and the 3D BBox $B_{0} \in \mathbb{R}^{7}$ in the initial frame. This can be described as
%(containing center location $\in \mathbb{R}^{3}$, box size $\in \mathbb{R}^3$ and rotation degree $\mathbb{R}^{1}$)

\begin{equation}
    (\widetilde{P}^{o}_{k+1}, \widetilde{B}_{k+1}) = \mathcal{F}(P^{s}_{k+1}, \widetilde{P}^{o}_{k})\,,
    \label{eq:trackDefine}
\end{equation}
where $\widetilde{P}^{o}_{k}$, $\widetilde{P}^{o}_{k+1}$ and $\widetilde{B}_{k+1}$ are the predicted target point cloud and 3D BBox in frame $k$ and $k+1$, respectively. By referring to P2B~\cite{qi_p2b_2020} and its follow-ups~\cite{fang_3d-siamrpn_2020, hui_3d_2021, hui_3d_2022, shan_ptt_2021, wang_mlvsnet_2021, zheng_box-aware_2021}, we summarize the typically 3D SOT loss as
\begin{equation}
    \mathcal{L} = \rho_{1} \cdot \mathcal{L}_{cla} + \rho_{2} \cdot \mathcal{L}_{prop} + \rho_{3} \cdot \mathcal{L}_{reg} + \rho_{4} \cdot \mathcal{L}_{box}\,,
    \label{eq:SOT_Loss}
\end{equation}
where $\rho$ is the manually-tuned hyperparameter, $\mathcal{L}_{cla}$, $\mathcal{L}_{prop}$, $\mathcal{L}_{reg}$ and $\mathcal{L}_{box}$ are the losses for foreground-background classification, confidences for the BBox proposals, voting offsets of the seed points, and offsets of the BBox proposals, respectively.

To address this task, we propose MixCycle, a novel semi-supervised framework for 3D SOT. Illustrated in \cref{fig:pipline}, MixCycle relies on a SOTMixup data augmentation strategy to tackle data sparsity and diversity (\cref{sec:SOTMixup}). Further, it utilizes self and forward-backward cycle consistencies as sources of supervision to cover the object appearance and motion variation~(\cref{sec:cycle_tracking}). Additionally, we apply rigid transformations to ground truth (GT) labels to generate search areas in unlabeled data~(\cref{sec:Implementation_Details}).

\subsection{SOTMixup}
\label{sec:SOTMixup}
Inspired by the great success of \emph{mixup}~\cite{zhang_mixup_2018}, we develop SOTMixup to supply diverse training samples and deal with the point cloud diversity problem, providing a solution for the \emph{mixup} application in binary classification in SOT task. With two image samples $(I_A,I_B)$, \emph{mixup} can be simply describe as creating an image %$I^{m}_i = \lambda I_i + (1-\lambda) I_j$ and $y^{m}_i = \lambda y_i + (1-\lambda) y_j$,
\begin{equation}
    I^{m}_A = \lambda I_A + (1-\lambda) I_B\,,
    \label{eq:mixup}
\end{equation}
\begin{equation}
    y^{m}_A = \lambda y_A + (1-\lambda) y_B\,,
    \label{eq:mixupLabel}
\end{equation}
where $\lambda \in [0,1]$ is the mixing rate, and $(y_A,y_B)$ are the image labels. Typically,  $\lambda$ follows a Beta distribution $\beta (\eta, \eta)$. A muti-class loss is then calculated as
\begin{equation}
    \mathcal{L}_{muti\_cal} = \lambda \cdot \mathcal{C}(\widetilde{y},y_A) + (1-\lambda) \cdot \mathcal{C}(\widetilde{y},y_B)\,,
    \label{eq:mixupLoss}
\end{equation}
where $\widetilde{y}$ is the predicted label, and $\mathcal{C}$ is the criterion (usually being the cross-entropy loss). Vanilla \emph{mixup} applies linear interpolation in aligned pixel space.
However, this operation is not suitable to unordered point clouds. Furthermore, one of the key challenges for the SOT task is to determine whether the proposal is positive or negative. Muti-class label interpolation approach in \emph{mixup} cannot be directly applied.
Specifically, given the search area PC $P_{A}$ and a random PC $P_{B}$ sampled in the training set, we employ \emph{mixup} to generate a frontground and background label pair $(\lambda \cdot y_{A}, (1-\lambda) \cdot y_{B})$. In practice, this label pair should be set to $(y_{A}, 0)$, as the points in $P_{B}$ mismatch the template and should be considered as background.

\begin{figure}[tbp]
  \centering
  \includegraphics[width=1.0\linewidth]{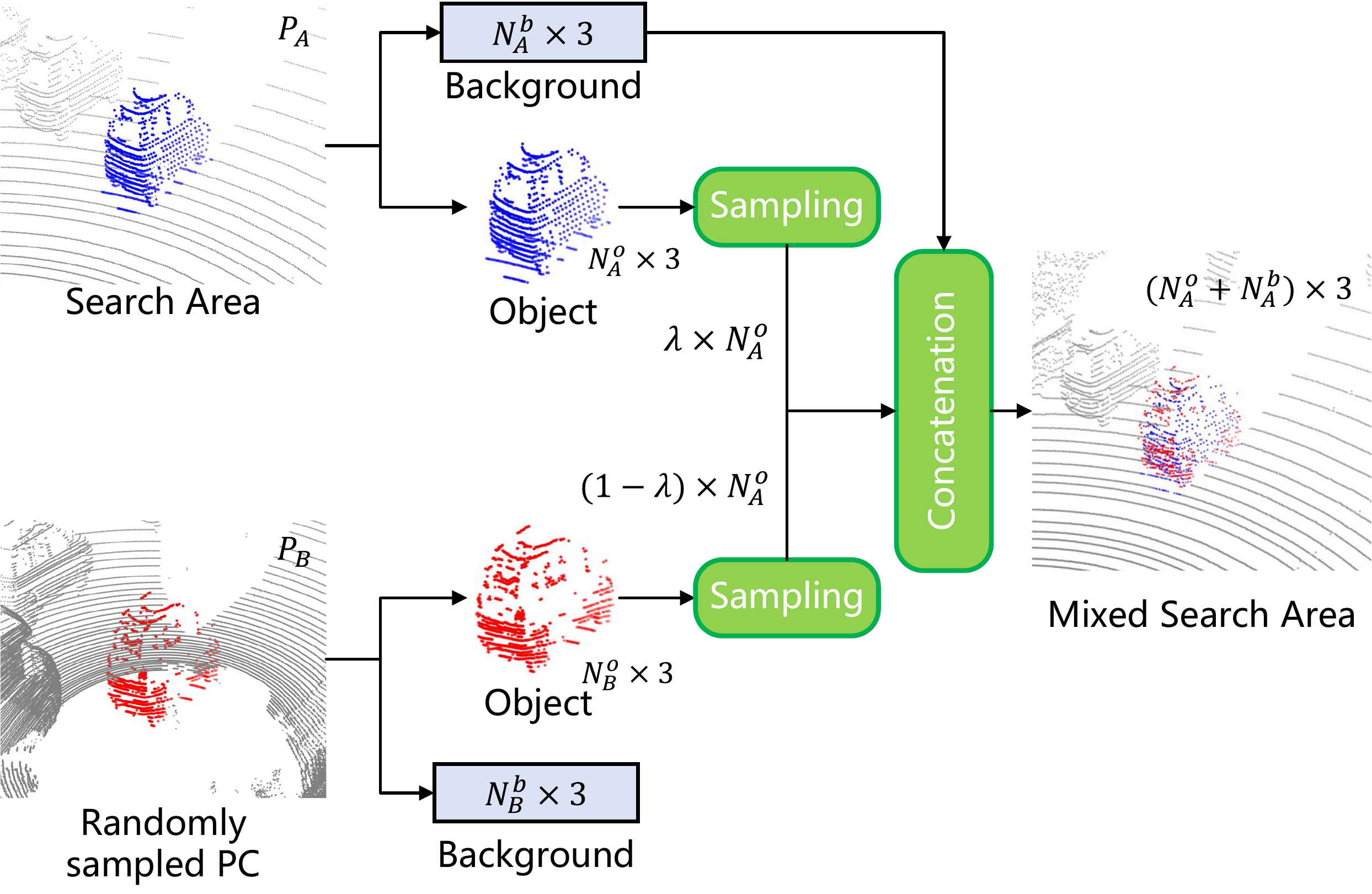}
    \caption{\textbf{SOTMixup.} First, the search area point cloud (PC) and a point cloud randomly sampled from the labeled training set are segmented into foreground and background, respectively. Second, a mixing rate $\lambda$ is applied to sample two object PCs. Finally, we concatenate the sampled object PCs and search area background to generate the mixed search area.}
   \label{fig:SOTMixup}
\end{figure}

We therefore develop a point cloud \emph{mixup} strategy for SOT based on the number of points, called SOTMixup. As shown in \cref{fig:SOTMixup}, SOTMixup generates new samples and minimizes the gap between the generated samples and the real sample distribution. Specifically, SOTMixup mixes a point cloud randomly sampled from the training set and the search area point cloud by sampling points using a mixing rate, without changing the total number of points in the search area. First, given point cloud pair $(P_{A},P_{B})$, corresponding binary classification labels $(y_A,y_B)$, and a mixing rate $\lambda$, we separate the backgrounds and object points in $P_{A}$ and $P_{B}$ and obtain  $(P^{b}_{A},P^{o}_{A},P^{b}_{B},P^{o}_{B} )$, where $P^{o}_{A} \in \mathbb{R}^{N^{o}_{A} \times 3}$ and $P^{o}_{B} \in \mathbb{R}^{N^{o}_{B} \times 3}$. 
Second, we generate $\hat{P^{o}_{A}}$ and $\hat{P^{o}_{B}}$ by randomly sampling $\lambda \times N^{o}_{A}$ and $(1-\lambda) \times N^{o}_{A}$ points from $P^{o}_{A}$ and $P^{o}_{B}$, respectively.
We then perform SOTMixup as
\begin{equation}
    P^{m}_{A} =  P^{b}_{A} + \hat{P^{o}_{A}} + \hat{P^{o}_{B}}\\,
    \label{eq:SOTMixup}
\end{equation}
where `$+$' represents the concatenation operation.

Usually, we consider the distance between the predicted object center and the ground truth to be positive if it is less than $0.3$ meters, and negative if it is greater than $0.6$ meters. The binary cross entropy loss for regression and foreground classification in SOTMixup can be written as
\begin{equation}
    \mathcal{L}_{prop\_mix} = -(\lambda \cdot y_{A} \cdot \log (s^{p}_{i}) + (1 - y_{A}) \cdot \log (1-s^{p}_i))\,,
    \label{eq:regressLoss_prop}
\end{equation}
\begin{equation}
    \mathcal{L}_{cla\_mix} = -(\lambda \cdot y_{A} \cdot \log (b^{p}_{j}) + (1 - y_{A}) \cdot \log (1-b^{p}_{j}))\,,
    \label{eq:regressLoss_cla}
\end{equation}
where $\mathcal{L}_{prop\_mix}$ and $\mathcal{L}_{cla\_mix}$ are the proposal confidence loss and foreground-background classification loss, respectively. $s^{p}_{i}$ is the confidence score of proposal $i$, and $b^{p}_{j}$ is the predicted foreground probability of point $j$ in search area $P^{s}$. We replace losses $\mathcal{L}_{prop}$ and $\mathcal{L}_{cla}$ with $\mathcal{L}_{prop\_mix} = \lambda \cdot \mathcal{L}_{prop}$ and $\mathcal{L}_{cla\_mix} = \lambda \cdot \mathcal{L}_{cla}$. SOTMixup applies a loss weight $\lambda$ to the positive proposals and foreground points, but does not change the loss weight of the negative proposals and background points. 
%We encourage the tracker to correctly predict the location of the object generated by SOTMixup and 
We reduce the loss penalty on the positive sample prediction scores to lessen the influence on the appearance matching ability of the tracker. We leave the loss weight unchanged for the negative samples. Because we intend the trackers to predict the motion offset of the object even if the object point cloud in the search area has dramatically changed.

\subsection{Cycle Tracking}
\label{sec:cycle_tracking}

%-------------------------------------------------------------------------------------

\noindent \textbf{Self Tracking Cycle.} In contrast with existing 2D cycle trackers~\cite{wang_unsupervised_2019,wang_unsupervised_2021,zheng_learning_2021,yuan_self-supervised_2020}, which only consider forward-backward cycle consistency, we propose to create the self tracking cycle. The motivation is to leverage virtually infinite supervision information contained in the initial frame itself. To this end, we first randomly sample a search area point cloud $P^{s}_{r}$ and object 3D BBox $B_{r}$ from the labeled training set. SOTMixup is then applied to generate the mixed point cloud
\begin{equation}
    P^{m}_{0} = SOTMixup(P^{s}_{0},B_{0},P^{s}_{r},B_{r},\lambda)\,,
\end{equation}
where $\lambda \in [0,1]$ is the mixing rate. 
Inspired by SC3D~\cite{giancola_leveraging_2019}, we apply a random rigid transformation $\mathcal{T}$ to $\{P^{m}_{0},B_{0}\}$ and generate pseudo labels 
\begin{equation}
    (P^{mt}_{0},B^{t}_{0}) = \mathcal{T}(P^{m}_{0},B_{0}, \alpha)\,,
    \label{eq:transformation}
\end{equation}
where $P_{0}^{mt}$ is the search area PC generated by applying SOT\textbf{M}ixup and a rigid \textbf{t}ransformation on $P_{0}^{s}$ and $\alpha = (\Delta x, \Delta y, \Delta z, \Delta \theta)$ is a transformation parameter with a coordinate offset $(\Delta x, \Delta y, \Delta z)$ and a rotation degree $\Delta \theta$ around the up-axis. This creates a self tracking cycle
\begin{equation}
    (\widetilde{P}^{ot}_{0}, \widetilde{B}^{t}_{0}) = \mathcal{F}(P^{mt}_{0}, P^{o}_{0})\,,
    \label{eq:trackself}
\end{equation}
where $\widetilde{B}^{t}_{0}$ is the predicted result on $P^{mt}_{0}$ and $P_{0}^{o}$ is the object template PC cropped from $P_{0}^{s}$. We then calculate the self consistency loss $\mathcal{L}_{self}$ between $\widetilde{B}^{t}_{0}$ and $B^{t}_{0}$. $\mathcal{L}_{self}$ has the same setting with $\mathcal{L}$ of~\cref{eq:SOT_Loss} while corresponding loss with $\mathcal{L}_{cal\_mix}$ and $\mathcal{L}_{prop\_mix}$.

In the self tracking cycle, the loss weight can be automatically quantified by the mixing rate in SOTMixup.
This provides the tracker with simple training samples to make it converge faster in the early stage of training with a high mixing rate, and also allows us to improve the tracker's robustness to point cloud variations using a low mixing rate.

%-------------------------------------------------------------------------------------

\noindent \textbf{Forward-Backward Tracking Cycle.} In addition to self tracking cycles, we also use forward-backward consistency. Hence, we forward track the object in the given search area sequence $\mathbf{P^s}$, which can be written as
\begin{equation}
    (\widetilde{P}^{o}_{1},\widetilde{B}_{1}) = \mathcal{F}(P^{s}_{1},P^{o}_{0})\,,
    \label{eq:trackforward1}
\end{equation}
\begin{equation}
    (\widetilde{P}^{o}_{2},\widetilde{B}_{2}) = \mathcal{F}(P^{s}_{2},\widetilde{P}^{o}_{1})\,,
    \label{eq:trackforward2}
\end{equation}
where $\{\widetilde{P}^{o}_{1}, \widetilde{P}^{o}_{2}\}$ and $\{\widetilde{B}_{1}, \widetilde{B}_{0}\}$ are the predicted forward tracking object point clouds and 3D BBoxes of $P^{s}_{1}$ and $P^{s}_{2}$, respectively. Following this strategy lets us further predict $\{\widetilde{P}^{o}_{3},\dots , \widetilde{P}^{o}_{n}\}$ and $\{\widetilde{B}_{3},\dots , \widetilde{B}_{n}\}$ in $\{P^{s}_{3},\dots , P^{s}_{n}\}$.

Then, we reverse the tracking sequence and perform backward tracking while applying random rigid transformations. This can be expressed as
\begin{equation}
    (\widetilde{P}^{o\prime}_{1},\widetilde{B}^{\prime}_{1}) = \mathcal{F}(\mathcal{T}(P^{s}_{1},\widetilde{B}_{1},\alpha),\widetilde{P}^{o\prime }_{2})\,,
    \label{eq:trackbackward1}
\end{equation}
\begin{equation}
    (\widetilde{P}^{o\prime}_{0},\widetilde{B}^{\prime}_{0}) = \mathcal{F}(\mathcal{T}(P^{s}_{0},{B}_{0},\alpha),\widetilde{P}^{o\prime}_{1})\,,
    \label{eq:trackbackward2}
\end{equation}
where $\{\widetilde{P}^{o\prime}_{0},\widetilde{P}^{o\prime}_{1}, \widetilde{P}^{o\prime}_{2}\}$ and $\{\widetilde{B}^{\prime}_{1}, \widetilde{B}^{\prime}_{0}\}$ are the predicted backward tracking object point clouds and 3D BBoxes. We then measure the consistency losses $\mathcal{L}_{con1}$ and $\mathcal{L}_{con0}$ between $\widetilde{B}^{\prime}_{1}$ and $\widetilde{B}_{1}$, as well as between $\widetilde{B}^{\prime}_{0}$ and $B_{0}$. Similarly, we can measure the consistency losses $\{\mathcal{L}_{con2}, \dots, \mathcal{L}_{con(n-1)}\}$. $\mathcal{L}_{con}$ has the same setting with $\mathcal{L}$ of~\cref{eq:SOT_Loss}.

The forward-backward tracking cycle provides real and diverse motion consistency, leading trackers to learn the object's motion distribution between two neighboring frames.
Furthermore, the tracker's robustness is increased by training with a disturbed template generated by the template update strategy (\cref{sec:Implementation_Details}).

%-------------------------------------------------------------------------------------

%-------------------------------------------------------------------------

\subsection{Implementation Details}
\label{sec:Implementation_Details}
\noindent \textbf{Training \& Testing.} We train MixCycle using the SGD optimizer with a batch size of $48$ and an initial learning rate of $0.01$ with a decay rate of $5e-5$ at each epoch. All experiments are conducted using NVIDIA RTX-3090 GPUs. We set $n=2$ and only measure the self and forward-backward cycle consistency losses $\mathcal{L}_{self}$ and $\mathcal{L}_{con0}$ for $P^{s}_{0}$ due to the GPU memory limit. At test time, we track the object frame by frame in a point cloud sequence with labels in the first frame. For both training and testing, our default setting for the template update strategy is to merge the target in the first frame with the predicted previous result.

\noindent \textbf{Input \& Data Augmentation.} Our MixCycle takes three frames $f$, $f+1$ and $f+2$ as input. For the initial frame $f$, we transform the BBox and point clouds to the object coordinate system. For the other frames in the tracking cycle, we transform the BBox and point clouds to the predicted object coordinate system in the last frame. We assume that the motion of the objects across neighboring frames is not significant. We apply random rigid transformations to BBoxes in labeled frames and use them to crop out search areas in the neighboring frames. Only the area within $2$ meters around the reference object BBox is considered as input (search area) since we are only interested in the area where the target is expected to appear. The random rigid transformation parameter $\alpha$ is set to $(0.3,0.3,0.0,5.0^{\circ})$, and the $\beta$ distribution parameter is set to $\eta = 0.5$.

\noindent \textbf{Loss Function.} The loss function of MixCycle is defined as $\mathcal{L}_{MixCycle} = \gamma_{1} \mathcal{L}_{self} + \gamma_{2} \mathcal{L}_{con0}$ containing self tracking cycle losses and forward-backward tracking cycle losses. Each of our cycle losses is set according to the original loss setting of the tracker to which we apply MixCycle. Additionally, the corresponding losses are replaced with $\mathcal{L}_{prop\_mix}$ and $\mathcal{L}_{cla\_mix}$. We empirically set $\gamma_{1} = 1.0$ and $\gamma_{2} = 2.0$, as we expect the tracker to focus more on learning the motion distribution of the object.
%-----------------------------------------------------------------------------
\section{Experiments}
\label{sec:experiment}

%-------------------------------------------------------------------------

\begin{table*}[htbp]\scriptsize
  \centering
  \caption{Overall performance comparison between our MixCycle and the fully-supervised methods on the KITTI (left) and NuScenes (right) datasets, where the percentage of labels used for training is shown under the dataset names. Improvements based on the same tracker are shown in \textcolor[rgb]{ 0,  .69,  .314}{green}.  \textbf{Bold} and \underline{underline} denote the best and the second-best performance, respectively.}
    \begin{tabular}{cc|cccccc|cccccc}
    \toprule[1.5pt]
                            & Dataset & \multicolumn{6}{c|}{KITTI}                    & \multicolumn{6}{c}{NuScenes} \\
\cmidrule{3-14}          & Sampling Rate & \multicolumn{2}{c|}{1\%} & \multicolumn{2}{c|}{5\%} & \multicolumn{2}{c|}{10\%} & \multicolumn{2}{c|}{0.1\%} & \multicolumn{2}{c|}{0.5\%} & \multicolumn{2}{c}{1\%} \\
    \midrule
    \midrule
    \multirow{6}[2]{*}{\begin{sideways}Success\end{sideways}} & P2B~\cite{qi_p2b_2020} & 6.1   &       & 25.5  &       & 34.3  &       & 15.2  &       & 23.0  &       & 24.3  &  \\
          & MLVSNet~\cite{wang_mlvsnet_2021} & \underline{25.0}  &       & 35.5  &       & 36.6  &       & 21.5  &       & 29.9  &       & 34.0  &  \\
          & BAT~\cite{zheng_box-aware_2021} & 10.1  &       & 21.6  &       & 34.9  &       & 17.5  &       & 26.4  &       & 30.6  &  \\
          & Ours(P2B) & 20.3  & \textcolor[rgb]{ 0,  .69,  .314}{14.2↑} & 36.7  & \textcolor[rgb]{ 0,  .69,  .314}{11.2↑} & \underline{43.8}  & \textcolor[rgb]{ 0,  .69,  .314}{9.5↑} & 23.2  & \textcolor[rgb]{ 0,  .69,  .314}{7.9↑} & \underline{34.3}  & \textcolor[rgb]{ 0,  .69,  .314}{11.3↑} & 34.3  & \textcolor[rgb]{ 0,  .69,  .314}{10.0↑} \\
          & Ours(MLVSNet) & \textbf{32.4 } & \textcolor[rgb]{ 0,  .69,  .314}{7.4↑} & \underline{38.8}  & \textcolor[rgb]{ 0,  .69,  .314}{3.3↑} & 42.6  & \textcolor[rgb]{ 0,  .69,  .314}{6↑} & \textbf{31.4 } & \textcolor[rgb]{ 0,  .69,  .314}{9.9↑} & \textbf{34.5 } & \textcolor[rgb]{ 0,  .69,  .314}{4.7↑} & \textbf{41.9 } & \textcolor[rgb]{ 0,  .69,  .314}{7.9↑} \\
          & Ours(BAT) & 19.7  & \textcolor[rgb]{ 0,  .69,  .314}{9.6↑} & \textbf{42.2 } & \textcolor[rgb]{ 0,  .69,  .314}{20.6↑} & \textbf{46.2 } & \textcolor[rgb]{ 0,  .69,  .314}{11.3↑} & \underline{24.4}  & \textcolor[rgb]{ 0,  .69,  .314}{6.9↑} & 32.8  & \textcolor[rgb]{ 0,  .69,  .314}{6.4↑} & \underline{34.4}  & \textcolor[rgb]{ 0,  .69,  .314}{3.8↑} \\
    \midrule
    \midrule
    \multirow{6}[2]{*}{\begin{sideways}Precision\end{sideways}} & P2B~\cite{qi_p2b_2020} & 5.0   &       & 39.5  &       & 52.7  &       & 13.0  & \textcolor[rgb]{ 0,  .69,  .314}{} & 21.2  & \textcolor[rgb]{ 0,  .69,  .314}{} & 22.6  &  \\
          & MLVSNet~\cite{wang_mlvsnet_2021} & \underline{36.3}  &       & 53.2  &       & 54.7  &       & 19.5  & \textcolor[rgb]{ 0,  .69,  .314}{} & 30.4  & \textcolor[rgb]{ 0,  .69,  .314}{} & \underline{35.3}  &  \\
          & BAT~\cite{zheng_box-aware_2021} & 12.3  &       & 35.3  &       & 52.7  &       & 15.2  & \textcolor[rgb]{ 0,  .69,  .314}{} & 25.7  & \textcolor[rgb]{ 0,  .69,  .314}{} & 30.6  &  \\
          & Ours(P2B) & 33.4  & \textcolor[rgb]{ 0,  .69,  .314}{28.4↑} & 55.3  & \textcolor[rgb]{ 0,  .69,  .314}{15.8↑} & \underline{64.2}  & \textcolor[rgb]{ 0,  .69,  .314}{11.5↑} & 21.9  & \textcolor[rgb]{ 0,  .69,  .314}{8.9↑} & \underline{34.2}  & \textcolor[rgb]{ 0,  .69,  .314}{13↑} & 34.0  & \textcolor[rgb]{ 0,  .69,  .314}{11.4↑} \\
          & Ours(MLVSNet) & \textbf{49.2 } & \textcolor[rgb]{ 0,  .69,  .314}{12.9↑} & \underline{56.6}  & \textcolor[rgb]{ 0,  .69,  .314}{3.4↑} & 61.4  & \textcolor[rgb]{ 0,  .69,  .314}{6.7↑} & \textbf{31.1 } & \textcolor[rgb]{ 0,  .69,  .314}{11.6↑} & \textbf{35.2 } & \textcolor[rgb]{ 0,  .69,  .314}{4.8↑} & \textbf{43.6 } & \textcolor[rgb]{ 0,  .69,  .314}{8.3↑} \\
          & Ours(BAT) & 27.0  & \textcolor[rgb]{ 0,  .69,  .314}{14.7↑} & \textbf{62.3 } & \textcolor[rgb]{ 0,  .69,  .314}{27.0↑} & \textbf{67.8 } & \textcolor[rgb]{ 0,  .69,  .314}{15.1↑} & \underline{22.7}  & \textcolor[rgb]{ 0,  .69,  .314}{7.5↑} & 31.9  & \textcolor[rgb]{ 0,  .69,  .314}{6.2↑} & 34.1  & \textcolor[rgb]{ 0,  .69,  .314}{3.5↑} \\
    \bottomrule[1.5pt]
    \end{tabular}%
  \label{tab:SummerizeTable}%
\end{table*}%

%-------------------------------------------------------------------------

\begin{table*}[htbp]
  \centering
  \caption{Comparison of MixCycle against fully-supervised methods on each category. We train the models with $1\%$/$0.1\%$ sampling rate on KITTI/NuScenes. Improvements and decreases based on the same tracker are shown in \textcolor[rgb]{ 0,  .69,  .314}{green} and \textcolor[rgb]{ 1,  0,  0}{red}, respectively.}
  \resizebox{\linewidth}{!}{
    \begin{tabular}{cc|cccccccccc|cccccccccc}
    \toprule[1.5pt]
                    & Dataset & \multicolumn{9}{c}{KITTI(1\%)}                                        &       & \multicolumn{10}{c}{Nuscenes(0.1\%)} \\
        \cmidrule{3-22}
          & Category & \multicolumn{2}{c}{Car} & \multicolumn{2}{c}{Pedestrian} & \multicolumn{2}{c}{Van} & \multicolumn{2}{c}{Cyclist} & \multicolumn{2}{c|}{Mean} & \multicolumn{2}{c}{Car} & \multicolumn{2}{c}{Truck} & \multicolumn{2}{c}{Trailer} & \multicolumn{2}{c}{Bus} & \multicolumn{2}{c}{Mean} \\
          & Frame Number & \multicolumn{2}{c}{6424} & \multicolumn{2}{c}{6088} & \multicolumn{2}{c}{1248} & \multicolumn{2}{c}{308} & \multicolumn{2}{c|}{14068} & \multicolumn{2}{c}{64159} & \multicolumn{2}{c}{13587} & \multicolumn{2}{c}{3352} & \multicolumn{2}{c}{2953} & \multicolumn{2}{c}{84051} \\
    \midrule
    \midrule
    \multirow{6}[2]{*}{\begin{sideways}Success\end{sideways}} & P2B~\cite{qi_p2b_2020}   & 8.1   &       & 3.6   &       & 8.1   &       & 5.6   &       & 6.1   &       & 15.8  &       & 13.1  &       & 12.8  &       & 16.1  &       & 15.2  &  \\
          & MLVSNet~\cite{wang_mlvsnet_2021} & \underline{35.3}  &       & 15.2  &       & \underline{22.9}  &       & 12.8  &       & \underline{25.0}  &       & 21.0  &       & 25.2  &       & 22.5  &       & 13.5  &       & 21.5  &  \\
          & BAT~\cite{zheng_box-aware_2021}   & 16.7  &       & 3.8   &       & 7.2   &       & 6.8   &       & 10.1  &       & 17.5  &       & 17.8  &       & 20.4  &       & 14.4  &       & 17.5  &  \\
          & Ours(P2B) & 20.6  & \textcolor[rgb]{ 0,  .69,  .314}{12.5↑} & \textbf{22.8 } & \textcolor[rgb]{ 0,  .69,  .314}{19.2↑} & 8.0   & \textcolor[rgb]{ 1,  0,  0}{0.1↓} & 16.6  & \textcolor[rgb]{ 0,  .69,  .314}{11.0↑} & 20.3  & \textcolor[rgb]{ 0,  .69,  .314}{14.2↑} & 23.0  & \textcolor[rgb]{ 0,  .69,  .314}{7.2↑} & 25.2  & \textcolor[rgb]{ 0,  .69,  .314}{12.1↑} & 22.4  & \textcolor[rgb]{ 0,  .69,  .314}{9.6↑} & \underline{17.7}  & \textcolor[rgb]{ 0,  .69,  .314}{1.5↑} & 23.2  & \textcolor[rgb]{ 0,  .69,  .314}{7.9↑} \\
          & Ours(MLVSNet) & \textbf{43.8 } & \textcolor[rgb]{ 0,  .69,  .314}{8.5↑} & \underline{20.7}  & \textcolor[rgb]{ 0,  .69,  .314}{5.5↑} & \textbf{28.2 } & \textcolor[rgb]{ 0,  .69,  .314}{5.3↑} & \textbf{43.7 } & \textcolor[rgb]{ 0,  .69,  .314}{31.0↑} & \textbf{32.4 } & \textcolor[rgb]{ 0,  .69,  .314}{7.4↑} & \textbf{29.7 } & \textcolor[rgb]{ 0,  .69,  .314}{8.7↑} & \textbf{42.4 } & \textcolor[rgb]{ 0,  .69,  .314}{17.3↑} & \textbf{31.3 } & \textcolor[rgb]{ 0,  .69,  .314}{8.9↑} & \textbf{19.2 } & \textcolor[rgb]{ 0,  .69,  .314}{5.7↑} & \textbf{31.4 } & \textcolor[rgb]{ 0,  .69,  .314}{9.9↑} \\
          & Ours(BAT) & 32.6  & \textcolor[rgb]{ 0,  .69,  .314}{15.9↑} & 6.1   & \textcolor[rgb]{ 0,  .69,  .314}{2.3↑} & 16.3  & \textcolor[rgb]{ 0,  .69,  .314}{9.2↑} & \underline{34.1}  & \textcolor[rgb]{ 0,  .69,  .314}{27.4↑} & 19.7  & \textcolor[rgb]{ 0,  .69,  .314}{9.6↑} & \underline{24.3}  & \textcolor[rgb]{ 0,  .69,  .314}{6.9↑} & \underline{26.9}  & \textcolor[rgb]{ 0,  .69,  .314}{9.1↑} & \underline{23.7}  & \textcolor[rgb]{ 0,  .69,  .314}{3.2↑} & 16.9  & \textcolor[rgb]{ 0,  .69,  .314}{2.5↑} & \underline{24.4}  & \textcolor[rgb]{ 0,  .69,  .314}{6.9↑} \\
    \midrule
    \midrule
    \multirow{6}[2]{*}{\begin{sideways}Precision\end{sideways}} & P2B~\cite{qi_p2b_2020}   & 7.4   & \textcolor[rgb]{ 0,  .69,  .314}{} & 2.2   & \textcolor[rgb]{ 0,  .69,  .314}{} & 6.1   & \textcolor[rgb]{ 0,  .69,  .314}{} & 4.4   & \textcolor[rgb]{ 0,  .69,  .314}{} & 5.0   & \textcolor[rgb]{ 0,  .69,  .314}{} & 14.5  & \textcolor[rgb]{ 0,  .69,  .314}{} & 8.2   & \textcolor[rgb]{ 0,  .69,  .314}{} & 6.8   & \textcolor[rgb]{ 0,  .69,  .314}{} & 8.4   & \textcolor[rgb]{ 0,  .69,  .314}{} & 13.0  & \textcolor[rgb]{ 0,  .69,  .314}{} \\
          & MLVSNet~\cite{wang_mlvsnet_2021} & \underline{46.5}  & \textcolor[rgb]{ 0,  .69,  .314}{} & 28.8  & \textcolor[rgb]{ 0,  .69,  .314}{} & \underline{25.4}  & \textcolor[rgb]{ 0,  .69,  .314}{} & 16.6  & \textcolor[rgb]{ 0,  .69,  .314}{} & \underline{36.3}  & \textcolor[rgb]{ 0,  .69,  .314}{} & 20.5  & \textcolor[rgb]{ 0,  .69,  .314}{} & 20.0  & \textcolor[rgb]{ 0,  .69,  .314}{} & 11.3  & \textcolor[rgb]{ 0,  .69,  .314}{} & 6.4   & \textcolor[rgb]{ 0,  .69,  .314}{} & 19.5  & \textcolor[rgb]{ 0,  .69,  .314}{} \\
          & BAT~\cite{zheng_box-aware_2021}   & 22.7  & \textcolor[rgb]{ 0,  .69,  .314}{} & 2.9   & \textcolor[rgb]{ 0,  .69,  .314}{} & 5.9   & \textcolor[rgb]{ 0,  .69,  .314}{} & 9.5   & \textcolor[rgb]{ 0,  .69,  .314}{} & 12.3  & \textcolor[rgb]{ 0,  .69,  .314}{} & 16.3  & \textcolor[rgb]{ 0,  .69,  .314}{} & 12.2  & \textcolor[rgb]{ 0,  .69,  .314}{} & 9.2   & \textcolor[rgb]{ 0,  .69,  .314}{} & \underline{12.2}  & \textcolor[rgb]{ 0,  .69,  .314}{} & 15.2  & \textcolor[rgb]{ 0,  .69,  .314}{} \\
          & Ours(P2B) & 30.0  & \textcolor[rgb]{ 0,  .69,  .314}{22.6↑} & \textbf{43.7 } & \textcolor[rgb]{ 0,  .69,  .314}{41.5↑} & 6.1   & 0.0   & 11.1  & \textcolor[rgb]{ 0,  .69,  .314}{6.7↑} & 33.4  & \textcolor[rgb]{ 0,  .69,  .314}{28.4↑} & 23.5  & \textcolor[rgb]{ 0,  .69,  .314}{9.0↑} & 18.9  & \textcolor[rgb]{ 0,  .69,  .314}{10.7↑} & 11.2  & \textcolor[rgb]{ 0,  .69,  .314}{4.4↑} & \textbf{14.0 } & \textcolor[rgb]{ 0,  .69,  .314}{5.6↑} & 21.9  & \textcolor[rgb]{ 0,  .69,  .314}{8.9↑} \\
          & Ours(MLVSNet) & \textbf{59.2 } & \textcolor[rgb]{ 0,  .69,  .314}{12.7↑} & \underline{40.7}  & \textcolor[rgb]{ 0,  .69,  .314}{11.9↑} & \textbf{31.1 } & \textcolor[rgb]{ 0,  .69,  .314}{5.7↑} & \textbf{79.0 } & \textcolor[rgb]{ 0,  .69,  .314}{62.4↑} & \textbf{49.2 } & \textcolor[rgb]{ 0,  .69,  .314}{12.8↑} & \textbf{31.1 } & \textcolor[rgb]{ 0,  .69,  .314}{10.6↑} & \textbf{38.6 } & \textcolor[rgb]{ 0,  .69,  .314}{18.6↑} & \textbf{19.5 } & \textcolor[rgb]{ 0,  .69,  .314}{8.1↑} & 11.5  & \textcolor[rgb]{ 0,  .69,  .314}{5.2↑} & \textbf{31.1 } & \textcolor[rgb]{ 0,  .69,  .314}{11.6↑} \\
          & Ours(BAT) & 43.9  & \textcolor[rgb]{ 0,  .69,  .314}{21.2↑} & 9.3   & \textcolor[rgb]{ 0,  .69,  .314}{6.4↑} & 19.2  & \textcolor[rgb]{ 0,  .69,  .314}{13.2↑} & \underline{57.3}  & \textcolor[rgb]{ 0,  .69,  .314}{47.8↑} & 27.0  & \textcolor[rgb]{ 0,  .69,  .314}{14.7↑} & \underline{24.1}  & \textcolor[rgb]{ 0,  .69,  .314}{7.8↑} & \underline{21.1}  & \textcolor[rgb]{ 0,  .69,  .314}{8.9↑} & \underline{13.8}  & \textcolor[rgb]{ 0,  .69,  .314}{4.6↑} & 9.7   & \textcolor[rgb]{ 1, 0, 0}{2.6↓} & \underline{22.7}  & \textcolor[rgb]{ 0,  .69,  .314}{7.5↑} \\
    \bottomrule[1.5pt]
    \end{tabular}%
    }
  \label{tab:categoryDetail}%
\end{table*}%

%-------------------------------------------------------------------------

\subsection{Datasets}
We evaluate our MixCycle on the challenging 3D visual tracking benchmarks of KITTI~\cite{geiger_are_2012}, NuScenes~\cite{caesar_nuscenes_2019} and Waymo~\cite{sun_scalability_2020} for semi-supervised 3D single object tracking. Semi-supervision labels are generated by applying random sampling to the training set.

The KITTI tracking dataset contains $21$ training sequences and $29$ test sequences with $8$ types of objects. Following previous works~\cite{giancola_leveraging_2019,qi_p2b_2020,zheng_box-aware_2021,wang_mlvsnet_2021}, we split the training set into training/validation/testing: Sequences $0$-$16$ are used for training, $17$-$18$ for validation, and $19$-$20$ for testing. The NuScenes dataset contains $1000$ scenes and annotations for $23$ object classes with accurate 3D BBoxes. NuScenes is officially divided into $700/150/150$ scenes for training/validation/testing. Following the setting in~\cite{zheng_box-aware_2021}, we train our MixCycle on the subset ``training\_track'' of the training set, and test it on the validation set. Waymo includes $1150$ scenes, $798/202/150$ scenes for training/validation/testing. Following the setting in~\cite{zheng_beyond_2022}, we test trackers in the validation set. Compared to KITTI, NuScenes, and Waymo include larger data volumes and more complex scenarios.

\noindent \textbf{Evaluation Metrics.} In this paper, we use One Pass Evaluation (OPE)~\cite{wu_online_2013} to evaluate the Success (Succ.) and Precision (Prec.) of different methods. \emph{Success} is calculated as the overlap (Intersection Over Union, IoU) between the proposal BBox and the ground truth (GT) BBox. \emph{Precision} represents the AUC of distance error between the centers of two BBoxes from $0$ to $2$ meters.

%-------------------------------------------------------------------------

\subsection{Comparison with Fully-supervised Methods}

To the best of our knowledge,  no other 3D single object trackers work in a semi-supervision fashion. Therefore, we choose P2B~\cite{qi_p2b_2020}, the multi-level voting Siamese network (MLVSNet)~\cite{wang_mlvsnet_2021} and the box-aware tracker (BAT)~\cite{zheng_box-aware_2021} to validate our method by sharing the same network backbone. Note that BAT is the state-of-the-art (SOTA) method in appearance matching-based trackers, and we regard it as our upper bound.
The fully-supervised methods will be trained with labeled data, as their original approaches.
We train MixCycle in a semi-supervised way, which uses both labeled and unlabeled data. We do not evaluate the motion-based tracker~\cite{zheng_beyond_2022} since it requires $2$ consecutive labeled point clouds for training and is not suitable for our training set generation strategy.
We employ different sampling rates for the two datasets. The first reason is that we account for the different scales of the dataset. The second season is that the case of very limited labels is more practical in real applications, and we attempt to set the sampling rate as low as possible within the trainable range.

%-------------------------------------------------------------------------

\begin{table}[tbp]
  \centering
  \caption{Overall performance comparison on KITTI/NuScenes between our MixCycle with $10\%$/$1\%$ sampling rates and the fully-supervised methods with $100\%$ sampling rate.}
  \resizebox{\linewidth}{!}{
    \begin{tabular}{c|c|cc|cc}
    \toprule[1.5pt]
    Dataset & Method & \multicolumn{2}{c|}{Success} & \multicolumn{2}{c}{Precision} \\
    \midrule
    \midrule
    \multirow{3}[2]{*}{KITTI} & P2B(100\%)~\cite{qi_p2b_2020} vs Ours(10\%) & 42.4 vs 43.8 & \textcolor[rgb]{ 0,  .69,  .314}{1.4↑} & 60.0 vs 64.2 & \textcolor[rgb]{ 0,  .69,  .314}{4.2↑} \\
          & MLVSNet(100\%)~\cite{wang_mlvsnet_2021} vs Ours(10\%) & 45.7 vs 42.6 & \textcolor[rgb]{ 1,  0,  0}{3.1↓} & 66.6 vs 61.4 & \textcolor[rgb]{ 1,  0,  0}{5.2↓} \\
          & BAT(100\%)~\cite{zheng_box-aware_2021} vs Ours(10\%) & 51.2 vs 46.2 & \textcolor[rgb]{ 1,  0,  0}{5.0↓} & 72.8 vs 67.8 & \textcolor[rgb]{ 1,  0,  0}{5.0↓} \\
    \midrule
    \midrule
    \multirow{3}[2]{*}{NuScenes} & P2B(100\%)~\cite{qi_p2b_2020} vs Ours(1\%) & 39.7 vs 34.3 & \textcolor[rgb]{ 1,  0,  0}{5.4↓} & 42.2 vs 34.0 & \textcolor[rgb]{ 1,  0,  0}{8.2↓} \\
          & MLVSNet(100\%)~\cite{wang_mlvsnet_2021} vs Ours(1\%) & 45.7 vs 41.9 & \textcolor[rgb]{ 1,  0,  0}{3.8↓}     & 47.9 vs 43.6 & \textcolor[rgb]{ 1,  0,  0}{4.3↓} \\
          & BAT(100\%)~\cite{zheng_box-aware_2021} vs Ours(1\%) & 41.8 vs 34.4 & \textcolor[rgb]{ 1,  0,  0}{7.4↓} & 42.7 vs 34.1 & \textcolor[rgb]{ 1,  0,  0}{8.6↓} \\
    \bottomrule[1.5pt]
    \end{tabular}%
    }
  \label{tab:100compare}%
\end{table}%

\begin{figure*}[t]
  \centering
    % \includegraphics[width=0.88\linewidth]{iccv2023AuthorKit/kitti_result_2.pdf}
    % \caption{Visualization. \textbf{Car}: Extremely sparse cases.  \textbf{Cyclist}: complex environment cases. }
  \includegraphics[width=0.90\linewidth]{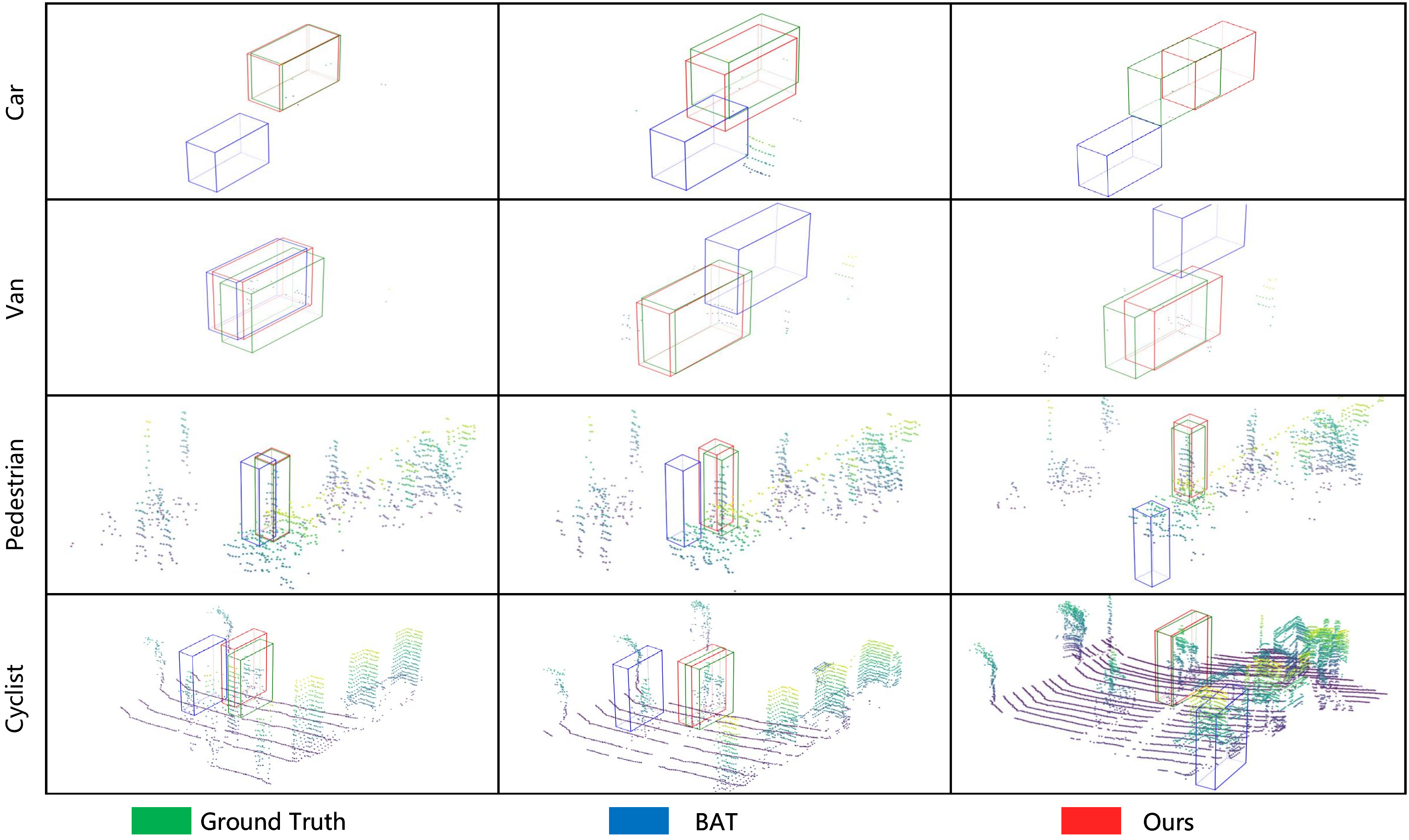}
    \caption{Visualization. \textbf{Car\&Van}: Extremely sparse cases. \textbf{Pedestrian}: Medium density cases. \textbf{Cyclist}: complex environment cases. }
  \label{fig:vis_result}
\end{figure*}

%-------------------------------------------------------------------------

\noindent \textbf{Results on KITTI.} \textbf{1)} We evaluate our MixCycle in $4$ categories (Car, Pedestrian, Van and Cyclist) and compare it using $3$ sampling rates: $1\%$, $5\%$ and $10\%$. As shown in \cref{tab:SummerizeTable}, our method outperforms the fully-supervised approaches under all sampling rates by a large margin. This confirms the high label efficiency of our proposed semi-supervised framework. 
% Table generated by Excel2LaTeX from sheet 'Sheet12'
\begin{table}[tbp]
  \centering
  \caption{Comparison of MixCycle against BAT on Car in KITTI for different sampling rates. `Improv.' denotes Improvement.}
  \resizebox{\linewidth}{!}{
    \begin{tabular}{cc|ccccccc}
    \toprule[1.5pt]
          & Sampling Rate & 1\%   & 5\%   & 10\%  & 30\%  & 50\%  & 70\%  & 100\% \\
    \midrule
    \midrule
    \multirow{3}[2]{*}{\begin{sideways}Succ.\end{sideways}} & BAT~\cite{zheng_box-aware_2021}   & 16.7  & 24.3  & 44.0  & 48.0  & 48.7  & 55.5  & 60.5  \\
          & Ours(BAT) & 32.6  & 49.2  & 55.2  & 56.2  & 56.6  & 60.9  & 64.7  \\
          & Improv. & \textcolor[rgb]{ 0,  .69,  .314}{15.9↑} & \textcolor[rgb]{ 0,  .69,  .314}{24.9↑} & \textcolor[rgb]{ 0,  .69,  .314}{11.2↑} & \textcolor[rgb]{ 0,  .69,  .314}{8.2↑} & \textcolor[rgb]{ 0,  .69,  .314}{7.9↑} & \textcolor[rgb]{ 0,  .69,  .314}{5.4↑} & \textcolor[rgb]{ 0,  .69,  .314}{4.2↑} \\
    \midrule
    \midrule
    \multirow{3}[2]{*}{\begin{sideways}Prec.\end{sideways}} & BAT~\cite{zheng_box-aware_2021}   & 22.7  & 34.8  & 57.3  & 63.1  & 65.3  & 69.5  & 77.7  \\
          & Ours(BAT) & 43.9  & 62.1  & 70.0  & 70.5  & 70.3  & 75.7  & 77.9  \\
          & Improv. & \textcolor[rgb]{ 0,  .69,  .314}{21.2↑} & \textcolor[rgb]{ 0,  .69,  .314}{27.3↑} & \textcolor[rgb]{ 0,  .69,  .314}{12.7↑} & \textcolor[rgb]{ 0,  .69,  .314}{7.4↑} & \textcolor[rgb]{ 0,  .69,  .314}{5.0↑} & \textcolor[rgb]{ 0,  .69,  .314}{6.2↑} & \textcolor[rgb]{ 0,  .69,  .314}{0.2↑} \\
    \bottomrule[1.5pt]
    \end{tabular}%
    }
  \label{tab:Car_diff_sample_rate}%
\end{table}%

The performance gap between our MixCycle and the fully-supervised P2B becomes larger as the proportion of labeled samples decreases. In particular, in the extreme case of $1\%$ labels usage, we achieve $\textbf{14.2\%}$ and $\textbf{28.4\%}$ improvement in success and precision, respectively, demonstrating the impact of our approach on the baseline method. Interestingly, our MixCycle based on the SOTA fully-supervised method BAT achieves the best results (($42.2\%$,$46.2\%$)/($62.3\%$,$67.8\%$) in succ./prec.) with $5\%$ and $10\%$ sampling rates, but the MLVSNet based MixCycle takes the first place ($32.4\%$/$49.2\%$ in succ./prec.) with $1\%$ sampling rate. We believe this is because the multi-scale approach in MLVSNet effectively enhances the robustness of its feature representation.
The BoxCloud proposed by BAT further strengthens the reliance on labels, leading to a degradation of BAT and MixCycle performance at $1\%$ sampling rate. \textbf{2)} We further present the test results on each category with $1\%$ sampling rate in \cref{tab:categoryDetail}. We achieve better performance in all categories, except Van on P2B. We assume this to be caused by the less labeled, huge, and moving fast feature of Van, which leads the tracker hard to predict a precise motion.
The improvements on Cyclist ($\textbf{31.0\%}$/$\textbf{62.4\%}$ and $\textbf{27.4\%}$/$\textbf{47.8\%}$ in succ./prec. for P2B and BAT, respectively) and 
Pedestrian ($\textbf{19.2\%}$/$\textbf{41.5\%}$ in succ./prec. on P2B) reveal the robustness of MixCycle to point cloud variations. As pedestrians and cyclists are usually considered to have the largest point cloud variations due to their small object sizes and the diversity of body motion. \textbf{3)} In \cref{tab:100compare}, we compare the performance between fully-supervised methods trained with $100\%$ labels and MixCycle trained with $10\%$ labels. Although our performance decreases slightly on MLVSNet and BAT, MixCycle still shows a remarkable result (43.8/64.2 in succ./prec.) on P2B. With only $10\%$ of the labels, MixCycle based on P2B outperforms the fully-supervised method ($1.4\%$/$4.2\%$ improvement in succ./prec.) using $100\%$ of the labels. This confirms the strong ability of our approach to leverage data information and highlights its promise for future developments. \textbf{4)} We provide the comparison of MixCycle against BAT on Car from KITTI with more sampling rates in \cref{tab:Car_diff_sample_rate}. Our Mixcycle not only achieves great improvements in low sampling rates, but also boosts the tracker's performance in the fully-supervised training (4.2/0.2 in succ./prec.). \textbf{5)} In \cref{fig:vis_result}, we compare MixCycle with BAT trained with $10\%$ labels. MixCycle achieves a better performance in both extremely sparse and complex point clouds.

\begin{table}[tbp]
  \caption{Comparison of MixCycle against BAT on Waymo. MixCycle(BAT) is trained only on KITTI with a $10\%$ sampling rate; BAT* represents BAT trained on Waymo using all the labels.}
  \resizebox{\linewidth}{!}{
    \begin{tabular}{c|cccccc}
    \toprule[1.5pt]
    % \multicolumn{1}{c|}{Category(Frame Number)} & \multicolumn{2}{c}{Vehicle(1057651)} & \multicolumn{2}{c}{Pedestrian(510533)} & \multicolumn{2}{c}{Mean(1568184)} \\
    \multicolumn{1}{c|}{Category} & \multicolumn{2}{c}{Vehicle} & \multicolumn{2}{c}{Pedestrian} & \multicolumn{2}{c}{Mean} \\
    \multicolumn{1}{c|}{Frame Number} & \multicolumn{2}{c}{1057651} & \multicolumn{2}{c}{510533} & \multicolumn{2}{c}{1568184} \\
    \multicolumn{1}{c|}{Metrics} & Succ. & \multicolumn{1}{c}{Prec.} & Succ. & \multicolumn{1}{c}{Prec.} & Succ. & \multicolumn{1}{c}{Prec.} \\
    \midrule
    \midrule
    BAT~\cite{zheng_box-aware_2021}   & 26.5  & 28.2  & 16.5  & 31.1  & 23.2  & 29.1  \\
    Ours(BAT) & \underline{31.1}  & \underline{33.5}  & \textbf{25.5} & \textbf{46.4} & \underline{29.3}  & \underline{37.7}  \\
    \midrule
    \multicolumn{1}{c|}{BAT*~\cite{zheng_box-aware_2021}} & \textbf{35.6} & \textbf{44.2} & \underline{22.1}  & \underline{36.8}  & \textbf{31.2 } & \textbf{41.8 } \\
    \bottomrule[1.5pt]
    \end{tabular}%
    }
  \label{tab:Waymo_Result}%
\end{table}%

\noindent \textbf{Results on NuScenes.} Following the setting in BAT~\cite{zheng_box-aware_2021}, we test our MixCycle in $4$ categories (Car, Truck, Trailer and Bus). The results of P2B~\cite{qi_p2b_2020}, MLVSNet~\cite{wang_mlvsnet_2021} and BAT~\cite{zheng_box-aware_2021} on NuScenes are provided by $M^2$-Track~\cite{zheng_beyond_2022} and BAT~\cite{zheng_box-aware_2021}. \textbf{1)} We use the published codes of the competitors to obtain results for each sampling rate. We compare them on $3$ sampling rates: $0.1\%$, $0.5\%$ and $1\%$, as NuScenes is larger than KITTI. As shown in \cref{tab:SummerizeTable}, MixCycle still outperforms the fully-supervised approaches under all sampling rates. \textbf{2)} Observing the individual categories in \cref{tab:categoryDetail} evidences that MixCycle yields a remarkable improvement on Truck ($\textbf{17.3\%}$/$\textbf{18.6\%}$ and $\textbf{12.1\%}$/$\textbf{10.7\%}$ in succ./prec. on MLVSNet and P2B, respectively) and Car ($8.7\%$/$10.6\%$ in succ./prec. on MLVSNet). However, MixCycle drops by $2.6\%$ in precision on Bus, which has fewer labels, a greater size, and high velocity. \textbf{3)} Moreover, in \cref{tab:100compare}, we compare the performance of MixCycle trained with $1\%$ labels and fully-supervised methods trained with $100\%$ labels. Our MixCycle based on MLVSNet surpasses the SOTA method BAT despite using significantly fewer labels. On such a challenging dataset with pervasive distractors and drastic appearance changes, our method exhibits even more competitive performance when using few labels.

\noindent\textbf{Result on Waymo.} Following the setting in~\cite{zheng_beyond_2022}, we test MixCycle in Vehicle and Pedestrian. We use the BAT backbone, and MixCycle(BAT) is only trained  on KITTI with a $10\%$ sampling rate. As shown in~\cref{tab:Waymo_Result}, MixCycle(BAT) outperforms BAT by \textbf{$6.1\%$} and \textbf{$8.6\%$} in terms of mean `succ.' and `prec.' values, respectively. 
{\textit{More impressively, our performance is close to the fully supervised BAT trained on Waymo ($31.2\%/41.75\%$ in succ./prec.~reported in~\cite{zheng_box-aware_2021}). }} To summarize, our MixCycle still delivers excellent results on large-scale datasets.

%-------------------------------------------------------------------------

\subsection{Analysis Experiments}

In this section, we extensively analyze MixCycle with a series of experiments. First, we study the effectiveness of each component in MixCycle. Second, we further analyze the influence of  forward-backward cycle step sizes. Finally, we compare the various application ways of SOTMixup. All the experiments are conducted on KITTI with $10\%$ sampling rate and with BAT as the backbone network, unless otherwise stated.

\noindent \textbf{Ablation Study.} We conduct experiments to analyze the effectiveness of different modules in MixCycle. First, we verify our assumption that appearance matching-based trackers can learn the object's motion distribution. As shown in \cref{tab:ablationstudy}, the cycle tracking framework yields better performance when using only forward-backward cycle than when using only self cycle ($3.5\%$/$5.4\%$ improvement in succ./prec.). Additionally, this supports the intuition that real and diverse motion information is helpful to appearance trackers. Furthermore, combining them boosts the results. Note that the random rigid transformation is necessary for self cycle, otherwise the GT BBox will always be fixed at the origin of the coordinate system. Second, we evaluate the effectiveness of random rigid transformation and SOTMixup in the framework. The performance grows significantly after applying SOTMixup ($5.1\%$/$6\%$ improvement in succ./prec.), demonstrating the importance of SOTMixup in semi-supervised tasks. Random rigid transformation plays a negative role in the cycle tracking framework but is practical in MixCycle. We conjecture this to be due to the missing target in the search area caused by random rigid transformation. This phenomenon may occur when the target moves rapidly and the model makes wrong predictions. Applying SOTMixup to the cycle tracking framework can significantly improve the model's tracking capabilities and address this issue.

\begin{table}[tbp]
  \centering
  \caption{Results of MixCycle when different modules are ablated. `Self', `F.-B. Cycle' and `Trans.' stand for self cycle, forward-backward cycle and random rigid transformation in the forward-backward cycle, respectively.}
  \resizebox{\linewidth}{!}{
    \begin{tabular}{cccc|cccc}
    \toprule[1.5pt]
    Self  & F.-B. Cycle & Trans.  & SOTMixup & \multicolumn{2}{c}{Success} & \multicolumn{2}{c}{Precision} \\
    \midrule
    \midrule
    \checkmark     & \multicolumn{1}{c}{} & \multicolumn{1}{c}{}     & \multicolumn{1}{c|}{} & 34.9  & \multicolumn{1}{p{3.065em}}{\textcolor[rgb]{ 1,  0,  0}{11.3↓}} & 52.7  & \multicolumn{1}{p{3.065em}}{\textcolor[rgb]{ 1,  0,  0}{15.1↓}} \\
    \multicolumn{1}{c}{} & \checkmark     & \multicolumn{1}{c}{} & \multicolumn{1}{c|}{} & 38.4  & \multicolumn{1}{p{3.065em}}{\textcolor[rgb]{ 1,  0,  0}{7.8↓}} & 58.1  & \multicolumn{1}{p{3.065em}}{\textcolor[rgb]{ 1,  0,  0}{9.7↓}} \\
    \checkmark     & \checkmark     & \multicolumn{1}{c}{} & \multicolumn{1}{c|}{} & 39.5  & \multicolumn{1}{p{3.065em}}{\textcolor[rgb]{ 1,  0,  0}{6.7↓}} & 59.2  & \multicolumn{1}{p{3.065em}}{\textcolor[rgb]{ 1,  0,  0}{8.6↓}} \\
    \checkmark     & \checkmark     & \checkmark     & \multicolumn{1}{c|}{} & 38.8  & \multicolumn{1}{p{3.065em}}{\textcolor[rgb]{ 1,  0,  0}{7.4↓}} & 59.3  & \multicolumn{1}{p{3.065em}}{\textcolor[rgb]{ 1,  0,  0}{8.5↓}} \\
    \checkmark     & \checkmark     & \multicolumn{1}{c}{} & \checkmark     & 44.6  & \multicolumn{1}{p{3.065em}}{\textcolor[rgb]{ 1,  0,  0}{1.6↓}} & 65.2  & \multicolumn{1}{p{3.065em}}{\textcolor[rgb]{ 1,  0,  0}{2.6↓}} \\
    \checkmark     & \checkmark     & \checkmark     & \checkmark     & \multicolumn{2}{c}{\textbf{46.2 }} & \multicolumn{2}{c}{\textbf{67.8 }} \\
    \bottomrule[1.5pt]
    \end{tabular}%
    }
  \label{tab:ablationstudy}%
\end{table}%

\begin{table}[tbp]\footnotesize
  \centering
  \caption{Analysis of the forward-backward cycle step size.}
    \begin{tabular}{c|cc}
    \toprule[1.5pt]
    \multicolumn{1}{c}{} & Success & Precision \\
    \midrule
    \midrule
    2 Steps & \multicolumn{1}{c}{45.8} & \multicolumn{1}{c}{66.6} \\
    3 Steps & \multicolumn{1}{c}{\textbf{46.2}} & \multicolumn{1}{c}{\textbf{67.8}} \\
    Improvement & \multicolumn{1}{c}{\textcolor[rgb]{0,  .69,  .314}{0.4↑}} & \multicolumn{1}{c}{\textcolor[rgb]{ 0,  .69,  .314}{1.2↑}} \\
    \bottomrule[1.5pt]
    \end{tabular}%
  \label{tab:stepTabel}%
\end{table}%

\begin{table}[tbp]
  \centering
  \caption{Results of SOTMixup with different settings. `Template' and `Search Area' indicate we apply SOTMixup with different inputs. `Self' indicates we apply SOTMixup in the self cycle. `Backward' means we apply SOTMixup in backward tracking $P^{s}_{1}$ to $P^{s}_{0}$.}
  \resizebox{\linewidth}{!}{
    \begin{tabular}{cccc|cccc}
    \toprule[1.5pt]
    Self  & Backward & Template & Search Area & \multicolumn{2}{c}{Success} & \multicolumn{2}{c}{Precision} \\
    \midrule
    \midrule
          & \checkmark     &       & \checkmark     & 44.7  & \textcolor[rgb]{ 1,  0,  0}{1.5↓} & 64.7  & \textcolor[rgb]{ 1,  0,  0}{3.1↓} \\
    \checkmark     & \checkmark     &       & \checkmark     & 45.4  & \textcolor[rgb]{ 1,  0,  0}{0.8↓} & 67.6  & \textcolor[rgb]{ 1,  0,  0}{0.2↓} \\
    \checkmark     &       & \checkmark     &       & 42.3  & \textcolor[rgb]{ 1,  0,  0}{3.9↓} & 63    & \textcolor[rgb]{ 1,  0,  0}{4.8↓} \\
    \checkmark     &       & \checkmark     & \checkmark     & 44.5  & \textcolor[rgb]{ 1,  0,  0}{1.7↓} & 66.1  & \textcolor[rgb]{ 1,  0,  0}{1.7↓} \\
    \checkmark     &       &       & \checkmark     & \multicolumn{2}{c}{\textbf{46.2 }} & \multicolumn{2}{c}{\textbf{67.8 }} \\
    \bottomrule[1.5pt]
    \end{tabular}%
    }
  \label{tab:SOTMixup}%
\end{table}%

\noindent \textbf{Flexibility of MixCycle.} We further explore the effect of forward-backward tacking cycle step size. As shown in \cref{tab:stepTabel}, we conduct experiments with $2$ and $3$ step cycles, respectively. Compared to the 2 step cycle, 3 step cycle achieves a better performance in both success and precision. This experiment demonstrates the potential of MixCycle for further growth in step size. We believe that larger step sizes can provide a template point cloud disturbed in long sequence tracking, leading to improved model robustness. Furthermore, according to~\cite{wang_unsupervised_2021}, a larger step size more effectively penalizes inaccurate localization. 

\noindent \textbf{Influence of SOTMixup.} In \cref{tab:SOTMixup}, we compare SOTMixup with different settings. First, we analyze the effect of applying SOTMixup to different inputs (Template and Search Area) while only using it in self cycle. The performance drops when we apply it to the template and to both the template and search area. We consider this to be due to the mismatch with real tracking. As the template is usually accurate while the search area point cloud varies significantly. Note that we share the same $\lambda$ when taking SOTMixup in both the template and search area, and set $\lambda = 1$ in the SOTMixup loss. Second, we explore various ways of exploiting SOTMixup  in MixCycle. SOTMixup leads to performance degradation when we apply it to backward tracking. This is caused by the disturbance of the template point cloud in backward tracking, making the loss weights misaligned. This further proves that the loss weights provided by SOTMixup are reliable.

%-------------------------------------------------------------------------
\section{Conclusion} In this paper, we have presented the first semi-supervised framework, MixCycle, for 3D SOT. Its three main components, self tracking cycle, forward-backward tracking cycle and SOTMixup, have been proposed to achieve robustness to point cloud variations and percept object's motion distribution. Our experiments have demonstrated that MixCycle yields high label efficiency and outperforming fully-supervised approaches using scarce labels.

In the future, we plan to develop a more robust tracking network backbone for MixCycle, and thus further enhance its 3D SOT performance.

\noindent\textbf{Acknowledgments.} This work is supported in part by the National Natural Science Foundation of China (No. 62176242 and 62002295), and NWPU international cooperation and exchange promotion projects (No. 02100-23GH0501). Thank Haozhe Qi for the discussion and preliminary exploration.

\clearpage

\appendix

\section{Implementation Details}

\begin{figure*}[htb]
    \centering
    \includegraphics[width=1.0\linewidth]{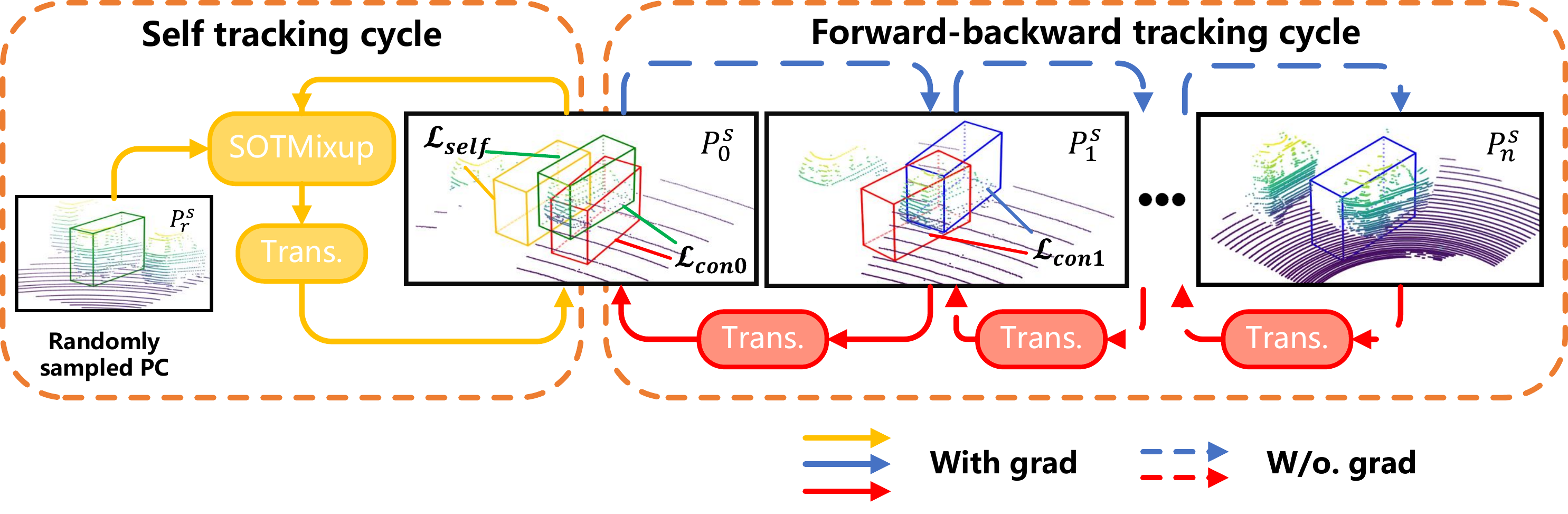}
    \caption{The framework of MixCycle. The gradient flow is represented by solid and dashed lines with arrows.}
    \label{fig:pipline_grad}
\end{figure*}

\begin{figure*}[htb]
    \centering
    \includegraphics[width=0.95\linewidth]{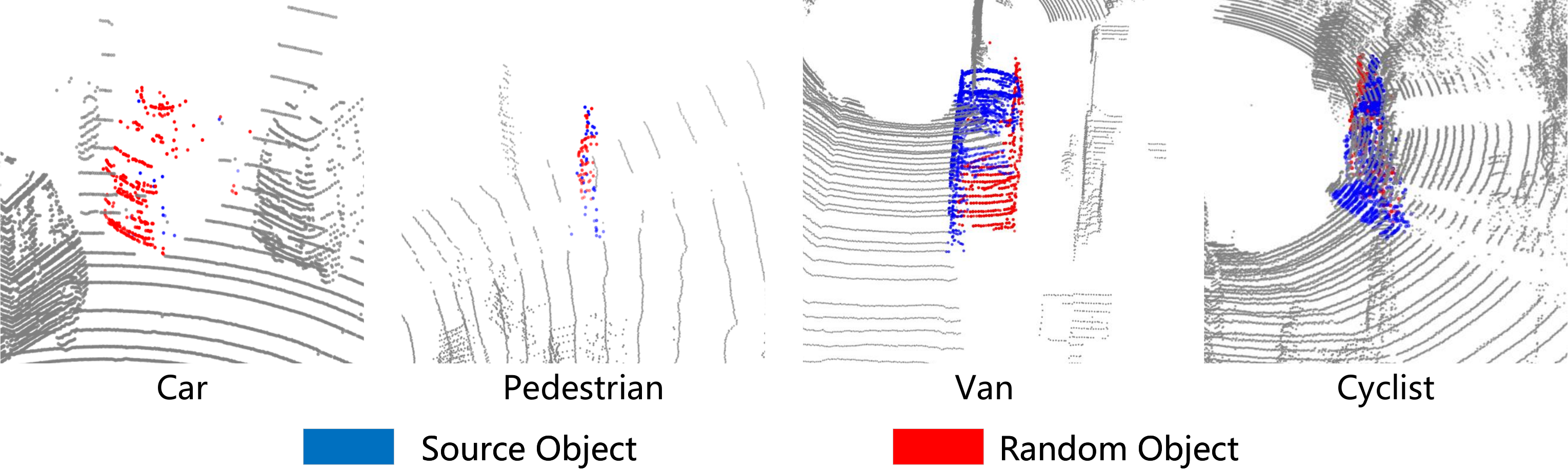}
    \caption{Visualization of SOTMixup.}
    \label{fig:SOTMixup_KITTI}
\end{figure*}

\begin{figure*}[htb]
    \centering
    \includegraphics[width=0.95\linewidth]{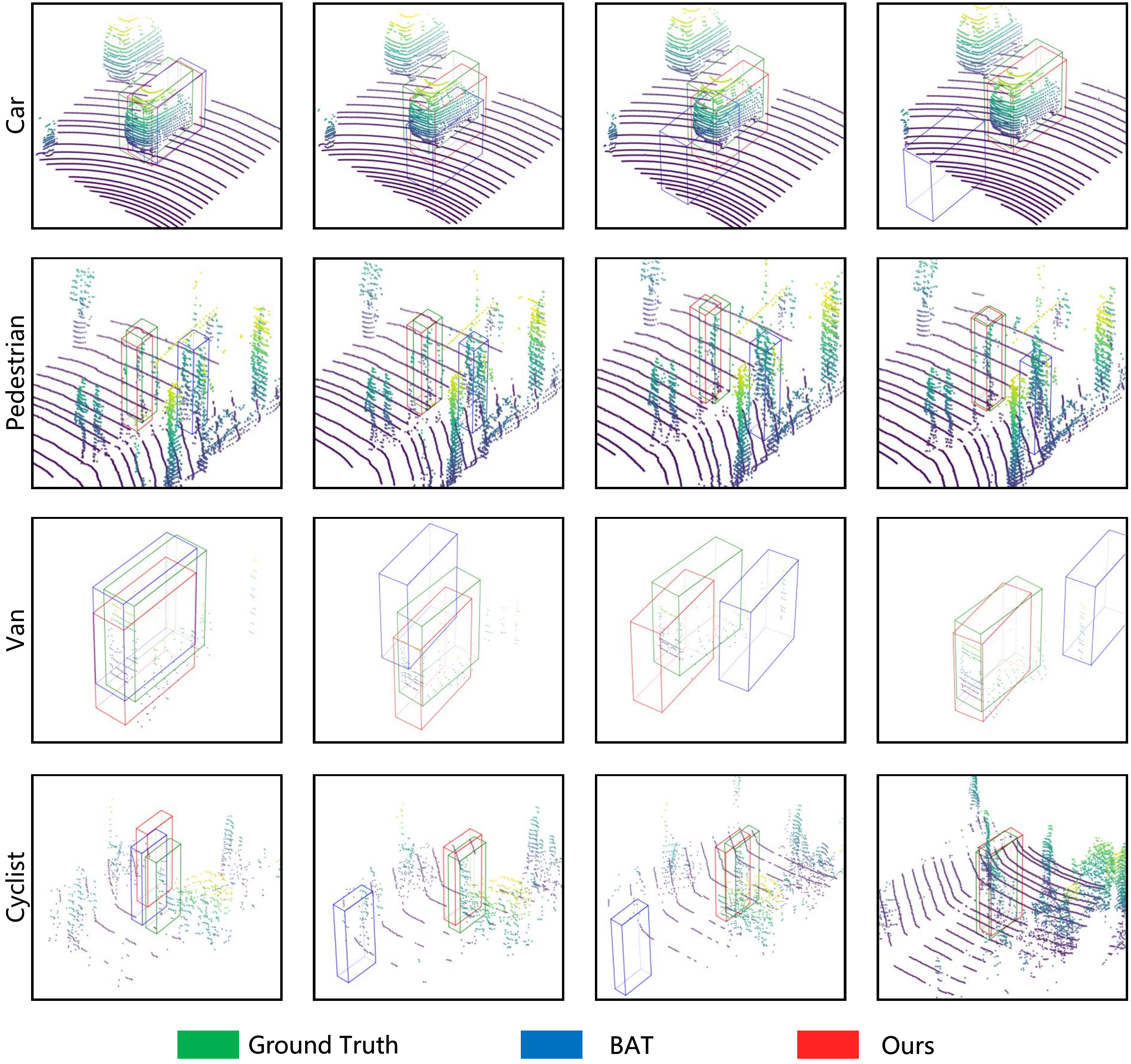}
    \caption{Visualization results. Our MixCycle and BAT are trained with $10\%$ labels on KITTI.}
    \label{fig:KITTI_vis}
\end{figure*}

\noindent \textbf{Framework Architecture.} The overall pipeline with the grad flow of MixCycle is shown in \cref{fig:pipline_grad}. Due to the limitation of non maximum suppression (NMS) on gradient back-propagation, we only calculate the gradients of the directly supervised parts. 

\noindent \textbf{SOTMixup.} Given the mix point cloud $P^{m}_{A} = P^{b}_{A} + \hat{P^{o}_{A}} + \hat{P^{o}_{B}}$ and Bounding-box $B_{A}$ in label $y_{A}$ at the SOTMixup, we only regard the points in $B_{A}$ as the foreground points. Specifically, the points in $\hat{P^{o}_{B}}$ are considered as background noise if they are outside the $B_{A}$. We believe that modifying the size of the tracking target is incompatible with real tracking.

% Introduce grad flow, forward details et. forward offset 
% Introduce training time & testing time

\section{More Analysis}

\noindent \textbf{Training \& Inference Time.} We compare MixCycle and fully-supervised methods~\cite{qi_p2b_2020,wang_mlvsnet_2021,zheng_box-aware_2021} in training time shown in \cref{tab:trainTime}. They are trained on Car in KITTI  with $10\%$ labels using $2$ NVIDIA RTX-3090 GPUs. Our MixCycle takes around $2.0 \sim 2.5$ times as long as the fully-supervised methods. The experiments reveal that MixCycle requires a longer training time, but it is still in an acceptable range. Hence, we could expect a faster and more robust tracking network backbone for MixCycle.

% Table generated by Excel2LaTeX from sheet 'Sheet2'
\begin{table}[htbp]
  \centering
  \caption{Training time comparison of MixCycle and fully-supervised methods on Car in KITTI with $10\%$ labels using $2$ NVIDIA RTX-3090 GPUs. Decreases based on the same tracker is shown in \textcolor[rgb]{1,0,0}{red}.}
    \begin{tabular}{c|cc}
    \toprule[1.5pt]
    Method & \multicolumn{2}{c}{Time} \\
    \midrule
    \midrule
    P2B~\cite{qi_p2b_2020}   & 1h22m &  \\
    MLVSNet~\cite{wang_mlvsnet_2021} & 1h47m &  \\
    BAT~\cite{zheng_box-aware_2021}   & 1h22m &  \\
    Ours(P2B) & 3h35m & \textcolor[rgb]{ 1,  0,  0}{2h13m↓} \\
    Ours(MLVSNet) & 3h32m & \textcolor[rgb]{ 1,  0,  0}{1h45m↓} \\
    Ours(BAT) & 3h40m & \textcolor[rgb]{ 1,  0,  0}{2h18m↓} \\
    \bottomrule[1.5pt]
    \end{tabular}%
  \label{tab:trainTime}%
\end{table}%

\begin{table*}[htbp]
  \centering
  \caption{Comparsion of MixCycle against fully-supervised methods on each category in KITTI. Improvements and decreases based on the same tracker are shown in \textcolor[rgb]{ 0,  .69,  .314}{green} and \textcolor[rgb]{1,0,0}{red}, respectively. \textbf{Bold} and \underline{underline} denote the best and the second-best performance, respectively.}
  \resizebox{\linewidth}{!}{
    \begin{tabular}{ccc|cccccccccc}
    \toprule[1.5pt]
          &       & Category & \multicolumn{2}{c}{Car} & \multicolumn{2}{c}{Pedestrian} & \multicolumn{2}{c}{Van} & \multicolumn{2}{c}{Cyclist} & \multicolumn{2}{c}{Mean} \\
          &       & Frame Number & \multicolumn{2}{c}{6424} & \multicolumn{2}{c}{6088} & \multicolumn{2}{c}{1248} & \multicolumn{2}{c}{308} & \multicolumn{2}{c}{14068} \\
    \midrule
    \midrule
    \multirow{18}[6]{*}{\begin{sideways}Success\end{sideways}} & \multirow{6}[2]{*}{1\%} & P2B~\cite{qi_p2b_2020}   & 8.11  &       & 3.61  &       & 8.10  &       & 5.60  &       & 6.11  &  \\
          &       & MLVSNet~\cite{wang_mlvsnet_2021} & \underline{35.27}  &       & 15.15  &       & \underline{22.94}  &       & 12.76  &       & \underline{24.98}  &  \\
          &       & BAT~\cite{zheng_box-aware_2021}   & 16.69  &       & 3.81  &       & 7.17  &       & 6.77  &       & 10.05  &  \\
          &       & Ours(P2B) & 20.56  & \textcolor[rgb]{ 0,  .69,  .314}{12.45↑} & \textbf{22.76 } & \textcolor[rgb]{ 0,  .69,  .314}{19.15↑} & 7.97  & \textcolor[rgb]{ 1,  0,  0}{0.13↓} & 16.62  & \textcolor[rgb]{ 0,  .69,  .314}{11.02↑} & 20.31  & \textcolor[rgb]{ 0,  .69,  .314}{14.20↑} \\
          &       & Ours(MLVSNet) & \textbf{43.75 } & \textcolor[rgb]{ 0,  .69,  .314}{8.48↑} & \underline{20.68}  & \textcolor[rgb]{ 0,  .69,  .314}{5.53↑} & \textbf{28.22 } & \textcolor[rgb]{ 0,  .69,  .314}{5.28↑} & \textbf{43.73 } & \textcolor[rgb]{ 0,  .69,  .314}{30.97↑} & \textbf{32.39 } & \textcolor[rgb]{ 0,  .69,  .314}{7.41↑} \\
          &       & Ours(BAT) & 32.63  & \textcolor[rgb]{ 0,  .69,  .314}{15.94↑} & 6.08  & \textcolor[rgb]{ 0,  .69,  .314}{2.27↑} & 16.33  & \textcolor[rgb]{ 0,  .69,  .314}{9.16↑} & \underline{34.12}  & \textcolor[rgb]{ 0,  .69,  .314}{27.35↑} & 19.73  & \textcolor[rgb]{ 0,  .69,  .314}{9.67↑} \\
\cmidrule{2-13}          & \multirow{6}[2]{*}{5\%} & P2B~\cite{qi_p2b_2020}   & 33.99  &       & 20.31  &       & 12.10  &       & 5.73  &       & 25.51  &  \\
          &       & MLVSNet~\cite{wang_mlvsnet_2021} & 43.50  &       & 28.09  &       & \underline{35.06}  &       & 19.77  &       & 35.55  &  \\
          &       & BAT~\cite{zheng_box-aware_2021}   & 24.30  &       & 21.00  &       & 13.17  &       & 13.25  &       & 21.62  &  \\
          &       & Ours(P2B) & 44.13  & \textcolor[rgb]{ 0,  .69,  .314}{10.14↑} & \underline{31.01}  & \textcolor[rgb]{ 0,  .69,  .314}{10.70↑} & 26.15  & \textcolor[rgb]{ 0,  .69,  .314}{14.05↑} & 36.77  & \textcolor[rgb]{ 0,  .69,  .314}{31.04↑} & 36.70  & \textcolor[rgb]{ 0,  .69,  .314}{11.19↑} \\
          &       & Ours(MLVSNet) & \textbf{52.44 } & \textcolor[rgb]{ 0,  .69,  .314}{8.94↑} & 24.04  & \textcolor[rgb]{ 1,  0,  0}{4.05↓} & \textbf{38.73 } & \textcolor[rgb]{ 0,  .69,  .314}{3.67↑} & \underline{46.54}  & \textcolor[rgb]{ 0,  .69,  .314}{26.77↑} & \underline{38.80}  & \textcolor[rgb]{ 0,  .69,  .314}{3.26↑} \\
          &       & Ours(BAT) & \underline{49.24}  & \textcolor[rgb]{ 0,  .69,  .314}{24.94↑} & \textbf{37.63 } & \textcolor[rgb]{ 0,  .69,  .314}{16.63↑} & 26.08  & \textcolor[rgb]{ 0,  .69,  .314}{12.91↑} & \textbf{50.08 } & \textcolor[rgb]{ 0,  .69,  .314}{36.83↑} & \textbf{42.18 } & \textcolor[rgb]{ 0,  .69,  .314}{20.56↑} \\
\cmidrule{2-13}          & \multirow{6}[2]{*}{10\%} & P2B~\cite{qi_p2b_2020}   & 41.94  &       & 30.63  &       & 19.61  &       & 7.37  &       & 34.31  &  \\
          &       & MLVSNet~\cite{wang_mlvsnet_2021} & 48.21  &       & 24.76  &       & 37.90  &       & 24.89  &       & 36.64  &  \\
          &       & BAT~\cite{zheng_box-aware_2021}   & 43.96  &       & 28.84  &       & 18.12  &       & 35.84  &       & 34.95  &  \\
          &       & Ours(P2B) & 45.82  & \textcolor[rgb]{ 0,  .69,  .314}{3.88↑} & \textbf{41.59 } & \textcolor[rgb]{ 0,  .69,  .314}{10.96↑} & \textbf{42.59 } & \textcolor[rgb]{ 0,  .69,  .314}{22.98↑} & \underline{52.14}  & \textcolor[rgb]{ 0,  .69,  .314}{44.77↑} & \underline{43.84}  & \textcolor[rgb]{ 0,  .69,  .314}{9.53↑} \\
          &       & Ours(MLVSNet) & \underline{54.08}  & \textcolor[rgb]{ 0,  .69,  .314}{5.87↑} & 30.39  & \textcolor[rgb]{ 0,  .69,  .314}{5.63↑} & \underline{41.29}  & \textcolor[rgb]{ 0,  .69,  .314}{3.39↑} & 49.95  & \textcolor[rgb]{ 0,  .69,  .314}{25.06↑} & 42.60  & \textcolor[rgb]{ 0,  .69,  .314}{5.97↑} \\
          &       & Ours(BAT) & \textbf{55.19 } & \textcolor[rgb]{ 0,  .69,  .314}{11.23↑} & \underline{38.62}  & \textcolor[rgb]{ 0,  .69,  .314}{9.78↑} & 34.92  & \textcolor[rgb]{ 0,  .69,  .314}{16.8↑} & \textbf{55.52 } & \textcolor[rgb]{ 0,  .69,  .314}{19.68↑} & \textbf{46.23 } & \textcolor[rgb]{ 0,  .69,  .314}{11.28↑} \\
    \midrule
    \midrule
    \multirow{18}[6]{*}{\begin{sideways}Precision\end{sideways}} & \multirow{6}[2]{*}{1\%} & P2B~\cite{qi_p2b_2020}   & 7.39  &       & 2.24  &       & 6.07  &       & 4.42  &       & 4.98  &  \\
          &       & MLVSNet~\cite{wang_mlvsnet_2021} & \underline{46.54}  &       & 28.80  &       & \underline{25.41}  &       & 16.62  &       & \underline{36.33}  &  \\
          &       & BAT~\cite{zheng_box-aware_2021}   & 22.66  &       & 2.92  &       & 5.94  &       & 9.54  &       & 12.35  &  \\
          &       & Ours(P2B) & 29.97  & \textcolor[rgb]{ 0,  .69,  .314}{22.58↑} & \textbf{43.73 } & \textcolor[rgb]{ 0,  .69,  .314}{41.49↑} & 6.08  & \textcolor[rgb]{ 0,  .69,  .314}{0.01↑} & 11.12  & \textcolor[rgb]{ 0,  .69,  .314}{6.70↑} & 33.39  & \textcolor[rgb]{ 0,  .69,  .314}{28.41↑} \\
          &       & Ours(MLVSNet) & \textbf{59.24 } & \textcolor[rgb]{ 0,  .69,  .314}{12.7↑} & \underline{40.72}  & \textcolor[rgb]{ 0,  .69,  .314}{11.92↑} & \textbf{31.08 } & \textcolor[rgb]{ 0,  .69,  .314}{5.67↑} & \textbf{79.03 } & \textcolor[rgb]{ 0,  .69,  .314}{62.41↑} & \textbf{49.16 } & \textcolor[rgb]{ 0,  .69,  .314}{12.83↑} \\
          &       & Ours(BAT) & 43.87  & \textcolor[rgb]{ 0,  .69,  .314}{21.21↑} & 9.32  & \textcolor[rgb]{ 0,  .69,  .314}{6.40↑} & 19.18  & \textcolor[rgb]{ 0,  .69,  .314}{13.24↑} & \underline{57.31}  & \textcolor[rgb]{ 0,  .69,  .314}{47.77↑} & 27.02  & \textcolor[rgb]{ 0,  .69,  .314}{14.67↑} \\
\cmidrule{2-13}          & \multirow{6}[2]{*}{5\%} & P2B~\cite{qi_p2b_2020}   & 45.99  &       & 40.26  &       & 10.82  &       & 5.43  &       & 39.50  &  \\
          &       & MLVSNet~\cite{wang_mlvsnet_2021} & 57.53  &       & 52.07  &       & \underline{42.30}  &       & 28.77  &       & 53.19  &  \\
          &       & BAT~\cite{zheng_box-aware_2021}   & 34.81  &       & 40.35  &       & 15.55  &       & 25.52  &       & 35.30  &  \\
          &       & Ours(P2B) & 56.94  & \textcolor[rgb]{ 0,  .69,  .314}{10.95↑} & \underline{58.04}  & \textcolor[rgb]{ 0,  .69,  .314}{17.78↑} & 30.92  & \textcolor[rgb]{ 0,  .69,  .314}{20.1↑} & 67.33  & \textcolor[rgb]{ 0,  .69,  .314}{61.90↑} & 55.34  & \textcolor[rgb]{ 0,  .69,  .314}{15.83↑} \\
          &       & Ours(MLVSNet) & \textbf{66.61 } & \textcolor[rgb]{ 0,  .69,  .314}{9.08↑} & 47.15  & \textcolor[rgb]{ 1,  0,  0}{4.92↓} & \textbf{45.26 } & \textcolor[rgb]{ 0,  .69,  .314}{2.96↑} & \underline{81.06}  & \textcolor[rgb]{ 0,  .69,  .314}{52.29↑} & \underline{56.61}  & \textcolor[rgb]{ 0,  .69,  .314}{3.42↑} \\
          &       & Ours(BAT) & \underline{62.07}  & \textcolor[rgb]{ 0,  .69,  .314}{27.26↑} & \textbf{68.05 } & \textcolor[rgb]{ 0,  .69,  .314}{27.70↑} & 30.81  & \textcolor[rgb]{ 0,  .69,  .314}{15.26↑} & \textbf{82.63 } & \textcolor[rgb]{ 0,  .69,  .314}{57.11↑} & \textbf{62.33 } & \textcolor[rgb]{ 0,  .69,  .314}{27.04↑} \\
\cmidrule{2-13}          & \multirow{6}[2]{*}{10\%} & P2B~\cite{qi_p2b_2020}   & 56.11  &       & 57.70  &       & 21.73  &       & 7.35  &       & 52.68  &  \\
          &       & MLVSNet~\cite{wang_mlvsnet_2021} & 63.63  &       & 48.31  &       & 44.65  &       & 35.08  &       & 54.69  &  \\
          &       & BAT~\cite{zheng_box-aware_2021}   & 57.25  &       & 56.08  &       & 21.48  &       & 19.69  &       & 52.75  &  \\
          &       & Ours(P2B) & 58.30  & \textcolor[rgb]{ 0,  .69,  .314}{2.19↑} & \textbf{72.05 } & \textcolor[rgb]{ 0,  .69,  .314}{14.35↑} & \textbf{51.83 } & \textcolor[rgb]{ 0,  .69,  .314}{30.1↑} & \underline{83.18}  & \textcolor[rgb]{ 0,  .69,  .314}{75.83↑} & \underline{64.22}  & \textcolor[rgb]{ 0,  .69,  .314}{11.54↑} \\
          &       & Ours(MLVSNet) & \underline{67.36}  & \textcolor[rgb]{ 0,  .69,  .314}{3.73↑} & 56.28  & \textcolor[rgb]{ 0,  .69,  .314}{7.97↑} & \underline{50.01}  & \textcolor[rgb]{ 0,  .69,  .314}{5.36↑} & 82.52  & \textcolor[rgb]{ 0,  .69,  .314}{47.44↑} & 61.36  & \textcolor[rgb]{ 0,  .69,  .314}{6.67↑} \\
          &       & Ours(BAT) & \textbf{70.02 } & \textcolor[rgb]{ 0,  .69,  .314}{12.77↑} & \underline{69.83}  & \textcolor[rgb]{ 0,  .69,  .314}{13.75↑} & 42.28  & \textcolor[rgb]{ 0,  .69,  .314}{20.8↑} & \textbf{85.37 } & \textcolor[rgb]{ 0,  .69,  .314}{65.68↑} & \textbf{67.81 } & \textcolor[rgb]{ 0,  .69,  .314}{15.06↑} \\
    \bottomrule[1.5pt]
    \end{tabular}%
    }
  \label{tab:result_KITTI_categories}%
\end{table*}%

% Table generated by Excel2LaTeX from sheet 'Sheet7'
\begin{table*}[htbp]
  \centering
  \caption{Comparsion of MixCycle against fully-supervised methods on each category in NuScenes.}
  \resizebox{\linewidth}{!}{
    \begin{tabular}{ccc|cccccccccc}
    \toprule[1.5pt]
          &       & Category & \multicolumn{2}{c}{Car} & \multicolumn{2}{c}{Truck} & \multicolumn{2}{c}{Trailer} & \multicolumn{2}{c}{Bus} & \multicolumn{2}{c}{Mean} \\
          &       & Frame Number & \multicolumn{2}{c}{64159} & \multicolumn{2}{c}{13587} & \multicolumn{2}{c}{3352} & \multicolumn{2}{c}{2953} & \multicolumn{2}{c}{84051} \\
    \midrule
    \midrule
    \multirow{18}[6]{*}{\begin{sideways}Success\end{sideways}} & \multirow{6}[2]{*}{0.1\%} & P2B~\cite{qi_p2b_2020}   & 15.77  &       & 13.09  &       & 12.81  &       & 16.12  &       & 15.23  &  \\
          &       & MLVSNet~\cite{wang_mlvsnet_2021} & 20.99  &       & 25.16  &       & 22.46  &       & 13.53  &       & 21.46  &  \\
          &       & BAT~\cite{zheng_box-aware_2021}   & 17.46  &       & 17.75  &       & 20.43  &       & 14.42  &       & 17.52  &  \\
          &       & Ours(P2B) & 23.01  & \textcolor[rgb]{ 0,  .69,  .314}{7.24↑} & 25.22  & \textcolor[rgb]{ 0,  .69,  .314}{12.13↑} & 22.37  & \textcolor[rgb]{ 0,  .69,  .314}{9.56↑} & \underline{17.65}  & \textcolor[rgb]{ 0,  .69,  .314}{1.53↑} & 23.15  & \textcolor[rgb]{ 0,  .69,  .314}{7.92↑} \\
          &       & Ours(MLVSNet) & \textbf{29.67 } & \textcolor[rgb]{ 0,  .69,  .314}{8.68↑} & \textbf{42.43 } & \textcolor[rgb]{ 0,  .69,  .314}{17.27↑} & \textbf{31.34 } & \textcolor[rgb]{ 0,  .69,  .314}{8.88↑} & \textbf{19.22 } & \textcolor[rgb]{ 0,  .69,  .314}{5.69↑} & \textbf{31.43 } & \textcolor[rgb]{ 0,  .69,  .314}{9.97↑} \\
          &       & Ours(BAT) & \underline{24.32}  & \textcolor[rgb]{ 0,  .69,  .314}{6.86↑} & \underline{26.88}  & \textcolor[rgb]{ 0,  .69,  .314}{9.13↑} & \underline{23.66}  & \textcolor[rgb]{ 0,  .69,  .314}{3.23↑} & 16.92  & \textcolor[rgb]{ 0,  .69,  .314}{2.50↑} & \underline{24.45}  & \textcolor[rgb]{ 0,  .69,  .314}{6.93↑} \\
\cmidrule{2-13}          & \multirow{6}[2]{*}{0.5\%} & P2B~\cite{qi_p2b_2020}   & 24.42  &       & 19.21  &       & 20.30  &       & 12.38  &       & 22.99  &  \\
          &       & MLVSNet~\cite{wang_mlvsnet_2021} & 29.82  &       & 32.25  &       & 27.40  &       & \underline{22.74}  &       & 29.87  &  \\
          &       & BAT~\cite{zheng_box-aware_2021}   & 27.71  &       & 22.85  &       & 25.48  &       & 15.44  &       & 26.40  &  \\
          &       & Ours(P2B) & \textbf{36.85 } & \textcolor[rgb]{ 0,  .69,  .314}{12.43↑} & 28.23  & \textcolor[rgb]{ 0,  .69,  .314}{9.02↑} & 21.75  & \textcolor[rgb]{ 0,  .69,  .314}{1.45↑} & 21.14  & \textcolor[rgb]{ 0,  .69,  .314}{8.76↑} & \underline{34.30}  & \textcolor[rgb]{ 0,  .69,  .314}{11.31↑} \\
          &       & Ours(MLVSNet) & 31.49  & \textcolor[rgb]{ 0,  .69,  .314}{1.67↑} & \textbf{46.75 } & \textcolor[rgb]{ 0,  .69,  .314}{14.50↑} & \textbf{48.49 } & \textcolor[rgb]{ 0,  .69,  .314}{21.09↑} & \textbf{28.47 } & \textcolor[rgb]{ 0,  .69,  .314}{5.73↑} & \textbf{34.53 } & \textcolor[rgb]{ 0,  .69,  .314}{4.66↑} \\
          &       & Ours(BAT) & \underline{32.20}  & \textcolor[rgb]{ 0,  .69,  .314}{4.49↑} & \underline{38.22}  & \textcolor[rgb]{ 0,  .69,  .314}{15.37↑} & \underline{31.04}  & \textcolor[rgb]{ 0,  .69,  .314}{5.56↑} & 21.82  & \textcolor[rgb]{ 0,  .69,  .314}{6.38↑} & 32.76  & \textcolor[rgb]{ 0,  .69,  .314}{6.36↑} \\
\cmidrule{2-13}          & \multirow{6}[2]{*}{1\%} & P2B~\cite{qi_p2b_2020}   & 23.95  &       & 27.83  &       & 25.84  &       & 14.57  &       & 24.32  &  \\
          &       & MLVSNet~\cite{wang_mlvsnet_2021} & 33.23  &       & \underline{39.08}  &       & 39.62  &       & 22.23  &       & 34.04  &  \\
          &       & BAT~\cite{zheng_box-aware_2021}   & 30.66  &       & 32.73  &       & 32.83  &       & 17.81  &       & 30.63  &  \\
          &       & Ours(P2B) & \underline{34.80}  & \textcolor[rgb]{ 0,  .69,  .314}{10.85↑} & 35.24  & \textcolor[rgb]{ 0,  .69,  .314}{7.41↑} & 30.40  & \textcolor[rgb]{ 0,  .69,  .314}{4.56↑} & 22.61  & \textcolor[rgb]{ 0,  .69,  .314}{8.04↑} & 33.43  & \textcolor[rgb]{ 0,  .69,  .314}{9.10↑} \\
          &       & Ours(MLVSNet) & \textbf{40.61 } & \textcolor[rgb]{ 0,  .69,  .314}{7.38↑} & \textbf{45.43 } & \textcolor[rgb]{ 0,  .69,  .314}{6.35↑} & \textbf{58.09 } & \textcolor[rgb]{ 0,  .69,  .314}{18.47↑} & \textbf{35.38 } & \textcolor[rgb]{ 0,  .69,  .314}{13.15↑} & \textbf{41.90 } & \textcolor[rgb]{ 0,  .69,  .314}{7.86↑} \\
          &       & Ours(BAT) & 33.72  & \textcolor[rgb]{ 0,  .69,  .314}{3.06↑} & 37.29  & \textcolor[rgb]{ 0,  .69,  .314}{4.56↑} & \underline{45.55}  & \textcolor[rgb]{ 0,  .69,  .314}{12.72↑} & \underline{24.26}  & \textcolor[rgb]{ 0,  .69,  .314}{6.45↑} & \underline{34.44}  & \textcolor[rgb]{ 0,  .69,  .314}{3.81↑} \\
    \midrule
    \midrule
    \multirow{18}[6]{*}{\begin{sideways}Precision\end{sideways}} & \multirow{6}[2]{*}{0.1\%} & P2B~\cite{qi_p2b_2020}   & 14.52  &       & 8.20  &       & 6.82  &       & 8.41  &       & 12.98  &  \\
          &       & MLVSNet~\cite{wang_mlvsnet_2021} & 20.45  &       & 19.97  &       & 11.31  &       & 6.35  &       & 19.51  &  \\
          &       & BAT~\cite{zheng_box-aware_2021}   & 16.31  &       & 12.16  &       & 9.19  &       & \underline{12.22}  &       & 15.21  &  \\
          &       & Ours(P2B) & 23.48  & \textcolor[rgb]{ 0,  .69,  .314}{8.96↑} & 18.88  & \textcolor[rgb]{ 0,  .69,  .314}{10.68↑} & 11.20  & \textcolor[rgb]{ 0,  .69,  .314}{4.38↑} & \textbf{13.99 } & \textcolor[rgb]{ 0,  .69,  .314}{5.58↑} & 21.91  & \textcolor[rgb]{ 0,  .69,  .314}{8.94↑} \\
          &       & Ours(MLVSNet) & \textbf{31.05 } & \textcolor[rgb]{ 0,  .69,  .314}{10.60↑} & \textbf{38.57 } & \textcolor[rgb]{ 0,  .69,  .314}{18.60↑} & \textbf{19.45 } & \textcolor[rgb]{ 0,  .69,  .314}{8.14↑} & 11.53  & \textcolor[rgb]{ 0,  .69,  .314}{5.18↑} & \textbf{31.12 } & \textcolor[rgb]{ 0,  .69,  .314}{11.60↑} \\
          &       & Ours(BAT) & \underline{24.10}  & \textcolor[rgb]{ 0,  .69,  .314}{7.79↑} & \underline{21.07}  & \textcolor[rgb]{ 0,  .69,  .314}{8.91↑} & \underline{13.81}  & \textcolor[rgb]{ 0,  .69,  .314}{4.62↑} & 9.67  & \textcolor[rgb]{ 1,  0,  0}{2.55↓} & \underline{22.69}  & \textcolor[rgb]{ 0,  .69,  .314}{7.48↑} \\
\cmidrule{2-13}          & \multirow{6}[2]{*}{0.5\%} & P2B~\cite{qi_p2b_2020}   & 24.28  &       & 12.32  &       & 11.08  &       & 6.98  &       & 21.21  &  \\
          &       & MLVSNet~\cite{wang_mlvsnet_2021} & 32.73  &       & 26.71  &       & 14.91  &       & \underline{15.35}  &       & 30.44  &  \\
          &       & BAT~\cite{zheng_box-aware_2021}   & 28.69  &       & 18.06  &       & 15.09  &       & 8.89  &       & 25.73  &  \\
          &       & Ours(P2B) & \textbf{39.22 } & \textcolor[rgb]{ 0,  .69,  .314}{14.94↑} & 20.79  & \textcolor[rgb]{ 0,  .69,  .314}{8.47↑} & 11.19  & \textcolor[rgb]{ 0,  .69,  .314}{0.11↑} & 13.27  & \textcolor[rgb]{ 0,  .69,  .314}{6.29↑} & \underline{34.21}  & \textcolor[rgb]{ 0,  .69,  .314}{13.00↑} \\
          &       & Ours(MLVSNet) & \underline{34.17}  & \textcolor[rgb]{ 0,  .69,  .314}{1.44↑} & \textbf{42.78 } & \textcolor[rgb]{ 0,  .69,  .314}{16.07↑} & \textbf{38.71 } & \textcolor[rgb]{ 0,  .69,  .314}{23.8↑} & \textbf{19.59 } & \textcolor[rgb]{ 0,  .69,  .314}{4.24↑} & \textbf{35.23 } & \textcolor[rgb]{ 0,  .69,  .314}{4.80↑} \\
          &       & Ours(BAT) & 33.35  & \textcolor[rgb]{ 0,  .69,  .314}{4.66↑} & \underline{32.21}  & \textcolor[rgb]{ 0,  .69,  .314}{14.15↑} & \underline{18.62}  & \textcolor[rgb]{ 0,  .69,  .314}{3.53↑} & 14.49  & \textcolor[rgb]{ 0,  .69,  .314}{5.60↑} & 31.92  & \textcolor[rgb]{ 0,  .69,  .314}{6.18↑} \\
\cmidrule{2-13}          & \multirow{6}[2]{*}{1\%} & P2B~\cite{qi_p2b_2020}   & 23.70  &       & 22.86  &       & 14.11  &       & 7.51  &       & 22.61  &  \\
          &       & MLVSNet~\cite{wang_mlvsnet_2021} & \underline{36.76}  &       & \underline{33.91}  &       & 29.60  &       & 15.41  &       & \underline{35.26}  &  \\
          &       & BAT~\cite{zheng_box-aware_2021}   & 32.47  &       & 28.36  &       & 20.42  &       & 11.19  &       & 30.58  &  \\
          &       & Ours(P2B) & 36.72  & \textcolor[rgb]{ 0,  .69,  .314}{13.02↑} & 29.15  & \textcolor[rgb]{ 0,  .69,  .314}{6.29↑} & 18.17  & \textcolor[rgb]{ 0,  .69,  .314}{4.06↑} & 14.78  & \textcolor[rgb]{ 0,  .69,  .314}{7.27↑} & 33.99  & \textcolor[rgb]{ 0,  .69,  .314}{11.38↑} \\
          &       & Ours(MLVSNet) & \textbf{45.07 } & \textcolor[rgb]{ 0,  .69,  .314}{8.31↑} & \textbf{40.17 } & \textcolor[rgb]{ 0,  .69,  .314}{6.26↑} & \textbf{46.28 } & \textcolor[rgb]{ 0,  .69,  .314}{16.68↑} & \textbf{25.01 } & \textcolor[rgb]{ 0,  .69,  .314}{9.60↑} & \textbf{43.62 } & \textcolor[rgb]{ 0,  .69,  .314}{8.36↑} \\
          &       & Ours(BAT) & 35.29  & \textcolor[rgb]{ 0,  .69,  .314}{2.82↑} & 32.30  & \textcolor[rgb]{ 0,  .69,  .314}{3.94↑} & \underline{32.63}  & \textcolor[rgb]{ 0,  .69,  .314}{12.21↑} & \underline{17.15}  & \textcolor[rgb]{ 0,  .69,  .314}{5.96↑} & 34.06  & \textcolor[rgb]{ 0,  .69,  .314}{3.49↑} \\
    \bottomrule[1.5pt]
    \end{tabular}%
    }
  \label{tab:result_NuScenes_categories}%
\end{table*}%

% Introduce performance on each categories

\noindent{\bf Frame Number of Cycle Tracking and Unlabeled Data Losses Balance.} \textbf{1)} Because of the limited memory of an NVIDIA RTX-3090 GPU, only a maximum of $2$ cycle consistencies among $3$ frames can be supervised. Therefore, we only present the losses for the self-supervised part.  \textbf{2)} For the one without labels part, we have made experiments to balance those losses. We try to supervise different consistencies in a two-stage training by supervising $\mathcal{L}_{self}$ and $\mathcal{L}_{con0}$ in stage $1$, and $\mathcal{L}_{self}$ and $\mathcal{L}_{con1}$ in stage $2$, based on BAT with a $10\%$ sampling rate on KITTI. Without SOTMixup, the cycle framework achieves $38.8$/$59.3$ and $41.0$/$60.6$ in Succ./Prec. in stage $1$ \& $2$, respectively. The performance drops in stage $2$ if we use SOTMixup. We conjecture this to be due to conflicts between the delicate losses set by SOTMixup in Self Cycle and the ambiguous losses in F.B. Cycle. We leave the design of a better training strategy for MixCycle as future work.

\noindent{\bf Fairness of Comparison with Fully-supervised Method.} Here we discuss fairness in the comparison experiments. The fully supervised method solely relies on labeled data, whereas our method utilizes both labeled and unlabeled data. \textbf{1)} The intention of our work is to reduce the effort in data annotation. While reducing the cost of collecting data is also important, we constitute a different research task on its own.
%and hopefully in the future we will be able to make it. 
\textbf{2)}  We refer to a semi-supervised 3D object detection method SESS's~\cite{zhao_sess_2020} experimental setting for comparison experiments. SESS directly reduces the usage of data of fully supervised methods for comparison experiments because no other method shares the same semi-supervised settings with it, which is very similar to our situation. \textbf{3)} We present the performance comparison using the same amount of data but with different label usage in the \cref{tab:100compare} and \cref{tab:Car_diff_sample_rate} of the paper. 

\noindent \textbf{Further Details.} We further demonstrate the test result on each category and sample rate on KITTI and Nuscenes shown in \cref{tab:result_KITTI_categories} and \cref{tab:result_NuScenes_categories}. We achieve great success on Cyclist. The maximum improvement on the Cyclist class is up to $\textbf{44.77\%}/\textbf{75.83\%}$ in success/precision based on P2B~\cite{qi_p2b_2020} with $10\%$ labels. For the most important class Car in KITTI and NuScenes, MixCycle also achieves a remarkable improvement in every sample rate. 
% Table generated by Excel2LaTeX from sheet 'Sheet10'

\section{Visualization}
\noindent \textbf{SOTMixup.} Our MixCycle leverages SOTMixup to supply diverse training samples. As shown in \cref{fig:SOTMixup_KITTI}, we present SOTMixup in a variety of categories. Our SOTMixup completes the point cloud of the occluded area in the Van in \cref{fig:SOTMixup_KITTI}, making the training samples more diverse. In the Car in \cref{fig:SOTMixup_KITTI}, SOTMixup almost removes the point cloud of the source object, allowing the trackers to regress the correct target center by learning the distribution of object motion in extreme cases.

\noindent \textbf{KITTI Results.} We present the visualization results of the comparison between Our MixCycle and BAT~\cite{zheng_box-aware_2021} with $10\%$ sample rate in \cref{fig:KITTI_vis}. The visualization results further validate the superiority of our approach in sparse and complex scenarios.

\clearpage
\clearpage

{\small
\bibliographystyle{ieee_fullname}
\bibliography{egbib}
}

\clearpage

\end{document}